\documentclass[10pt,twocolumn,letterpaper]{article}

\usepackage[pagenumbers]{cvpr}

\usepackage[dvipsnames]{xcolor}

\definecolor{cvprblue}{rgb}{0.21,0.49,0.74}
\usepackage[pagebackref,breaklinks,colorlinks,citecolor=cvprblue]{hyperref}

\usepackage{amsmath,amsfonts,bm}

\def\eqref#1{equation~\ref{#1}}

\def\1{\bm{1}}

\DeclareMathAlphabet{\mathsfit}{\encodingdefault}{\sfdefault}{m}{sl}
\SetMathAlphabet{\mathsfit}{bold}{\encodingdefault}{\sfdefault}{bx}{n}

\usepackage{hyperref}
\usepackage{url}
\usepackage{times}
\usepackage{soul}
\usepackage{url}
\usepackage[utf8]{inputenc}
\usepackage{graphicx}
\usepackage{amsmath}
\usepackage{amsthm}
\usepackage{booktabs}
\usepackage{algorithm}
\usepackage{algorithmic}
\usepackage[switch]{lineno}

\title{Benchmarking Zero-Shot Robustness of Multimodal Foundation Models: \\A Pilot Study}

\author{Chenguang Wang\textsuperscript{1}$^*$, Ruoxi Jia\textsuperscript{2}, Xin Liu\textsuperscript{3}, Dawn Song\textsuperscript{4}\\
\textsuperscript{1}Washington University in St. Louis, \textsuperscript{2}Virginia Tech, \textsuperscript{3}UC Davis, \textsuperscript{4}UC Berkeley\\
{\tt\small chenguangwang@wustl.edu, ruoxijia@vt.edu, xinliu@ucdavis.edu, dawnsong@berkeley.edu}}

\usepackage[utf8]{inputenc} 
\usepackage[T1]{fontenc}    
\usepackage{hyperref}       
\usepackage{url}            
\usepackage{amsfonts}       
\usepackage{nicefrac}       
\usepackage{microtype}      
\usepackage{xcolor}         

\usepackage{enumitem}
\usepackage{times}
\usepackage{graphicx}
\usepackage{latexsym}
\usepackage{hyperref}
\usepackage[hyphenbreaks]{breakurl}
\usepackage{url}
\usepackage{comment}
\usepackage{wrapfig}
\usepackage{graphics}
\usepackage{multirow}
\usepackage{colortbl}
\usepackage{makecell}
\usepackage{arydshln}
\usepackage{enumitem}
\usepackage{mdwlist}
\usepackage{float}
\usepackage[normalem]{ulem}
\useunder{\uline}{\ul}{}
\usepackage{subcaption}
\usepackage{wrapfig,lipsum,booktabs}

\usepackage[T1]{fontenc}

\usepackage[utf8]{inputenc}

\usepackage{microtype}

\usepackage[list=true]{subcaption}
\usepackage{amsmath,amssymb}

\def\best{\bf\cellcolor[gray]{0.85}}

\definecolor{TableRed}{HTML}{800000}
\definecolor{mediumelectricblue}{rgb}{0.01, 0.31, 0.59}

\mathchardef\mhyphen="2D 

\newcommand{\benchmark}{\textsc{RoZ}}

\newcommand{\ours}{(\textcolor{mediumelectricblue}{\small ours})}
\newcommand{\comm}[1]{}
\newcommand{\textspt}[1]{\texttt{#1}}

\begin{document}
\maketitle

\def\thefootnote{$^*$}\footnotetext{Corresponding author.}
\def\thefootnote{\arabic{footnote}}
\def\thefootnote{$\color{white} ^*$}\footnotetext{The code and datasets are available at \url{https://github.com/wang-research-lab/roz}.}
\def\thefootnote{\arabic{footnote}}

\begin{abstract}
    Pre-training image representations from the raw text about images enables zero-shot vision transfer to downstream tasks. Through pre-training on millions of samples collected from the internet, multimodal foundation models, such as CLIP, produce state-of-the-art zero-shot results that often reach competitiveness with fully supervised methods without the need for task-specific training. Besides the encouraging performance on classification accuracy, it is reported that these models close the robustness gap by matching the performance of supervised models trained on ImageNet under natural distribution shift. Because robustness is critical to real-world applications, especially safety-critical ones, in this paper, we present a comprehensive evaluation based on a large-scale robustness benchmark covering 7 natural, 3 synthetic distribution shifts, and 11 adversarial attacks. We use CLIP as a pilot study. We show that CLIP leads to a significant robustness drop compared to supervised ImageNet models on our benchmark, especially under synthetic distribution shift and adversarial attacks. Furthermore, data overlap analysis suggests that the observed robustness under natural distribution shifts could be attributed, at least in part, to data overlap. In summary, our evaluation shows a comprehensive evaluation of robustness is necessary; and there is a significant need to improve the robustness of zero-shot multimodal models.
\end{abstract}

\section{Introduction}

The common recipe of current state-of-the-art multimodal foundation models is the pre-training that learns representations from images and raw text. 
At test time, a standardized interface of natural language prompts enables task-agnostic architectures to zero-shot transfer to downstream datasets without the need for dataset-specific training or architecture modifications. For example, multimodal foundation models such as CLIP~\cite{radford2021learning} learn image representations on a pre-training dataset of hundreds of millions of samples collected from the web, and is competitive across many computer vision tasks zero-shot, even comparable to task-specific fully supervised methods.

Besides the superior performance, multimodal models have reportedly made similar breakthroughs in robustness. For example, as one of the pioneer models, zero-shot CLIP has closed the robustness gap by up to 75\% while matching the performance of a standard model trained on ImageNet.
However, the robustness is often tested on natural distribution shifts, which contain natural (or unmodified) images collected from the web. While investigating their robustness to natural distribution shifts is important, it remains unclear whether these models are robust to synthetic distrubution shifts, such as noise corruptions \cite{hendrycks2019benchmarking} and spatial transformations~\cite{barbu2019objectnet}, and adversarial examples~\cite{goodfellow2014explaining}. This is essential for these models to be deployed in safety-critical applications.

\begin{figure*}
\centering
\subcaptionbox{{Seven natural distribution shifts.}\label{fig:alldista}}{\includegraphics[width=0.48\textwidth]{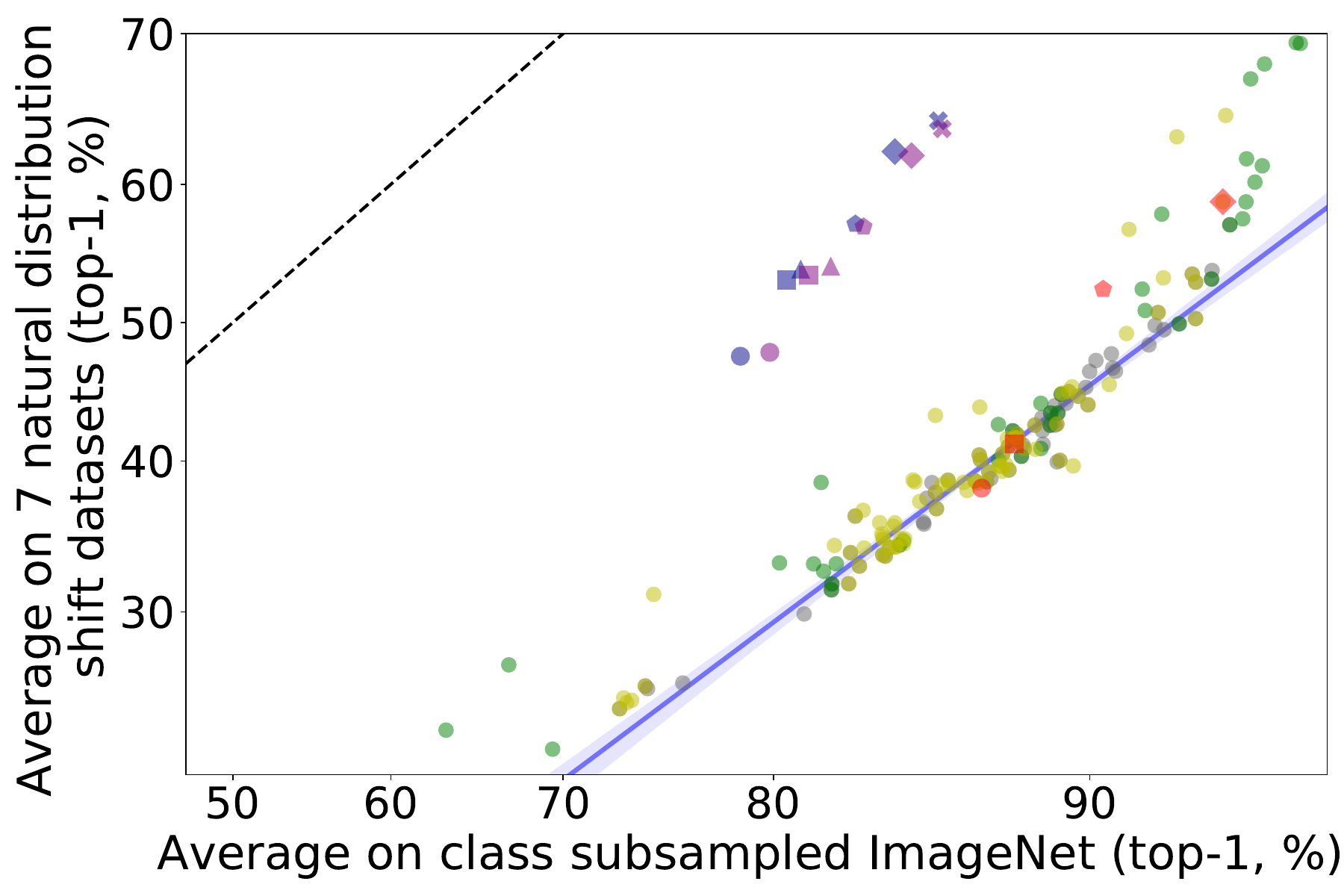}}%
\hspace{0.06in}
\subcaptionbox{{Robustness test sets except for the distribution shifts in (a).}\label{fig:alladvb}}{\includegraphics[width=0.48\textwidth]{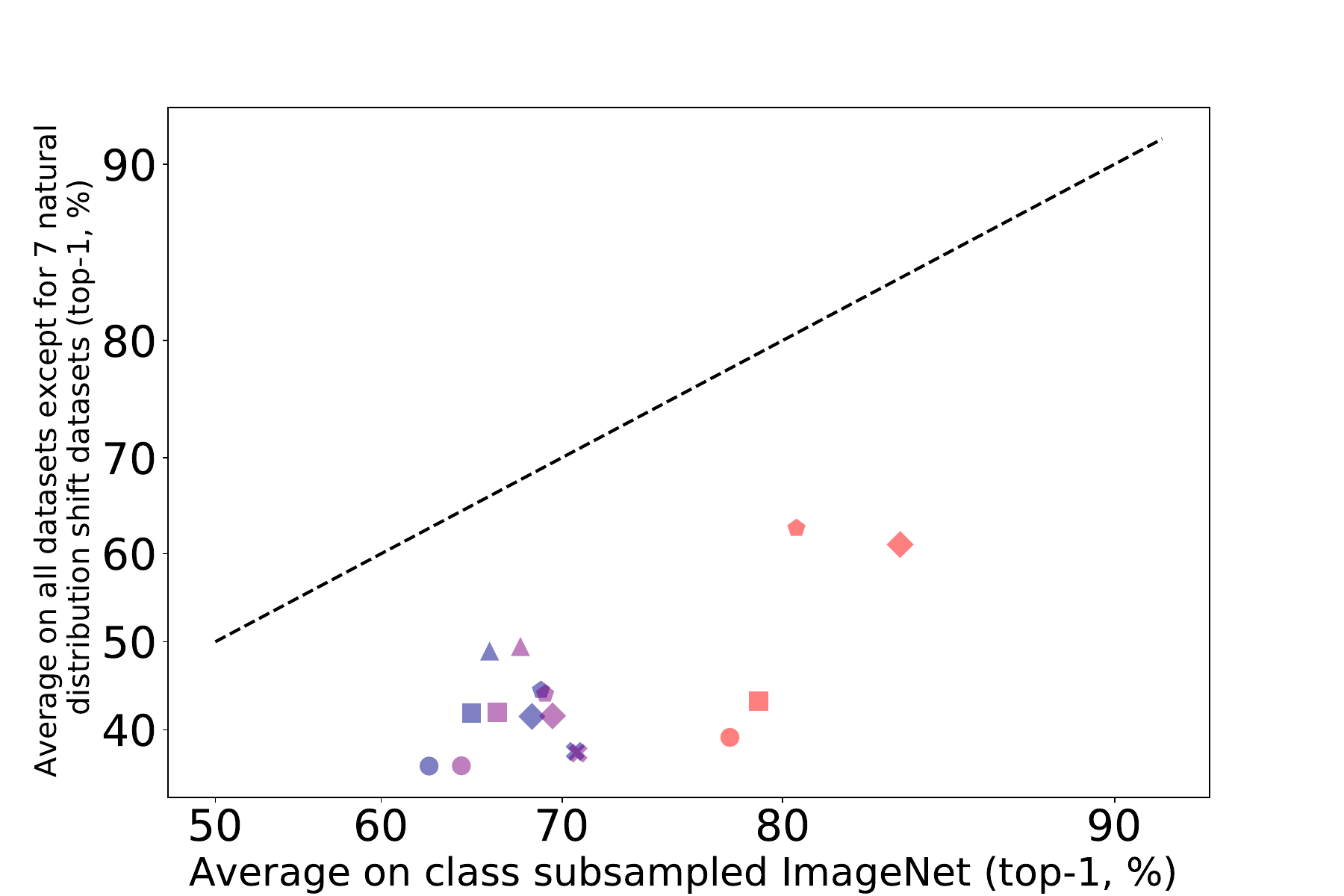}}%
\\
\includegraphics[width=0.95\linewidth]{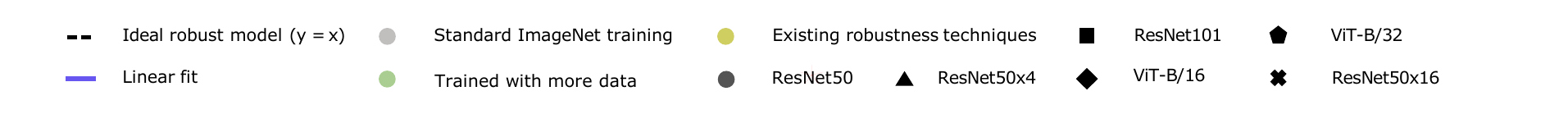}%
\vspace{-0.15in}
\caption{{Summary of results on our \benchmark\ benchmark. An ideal robust model (dashed line) performs equally well on the ImageNet distribution and other distributions. Multimodal models such as CLIP fail to improve robustness on test sets in (b) of our benchmark except for the test sets in (a). Red: standard ImageNet models. Blue: zero-shot CLIP models. Purple: CLIP-Auto models.}}
    \label{fig:introfig}
\end{figure*}

\begin{figure*}[t]
\centering
\subcaptionbox{{All datasets.}\label{fig:all}}{\includegraphics[width=0.48\textwidth]{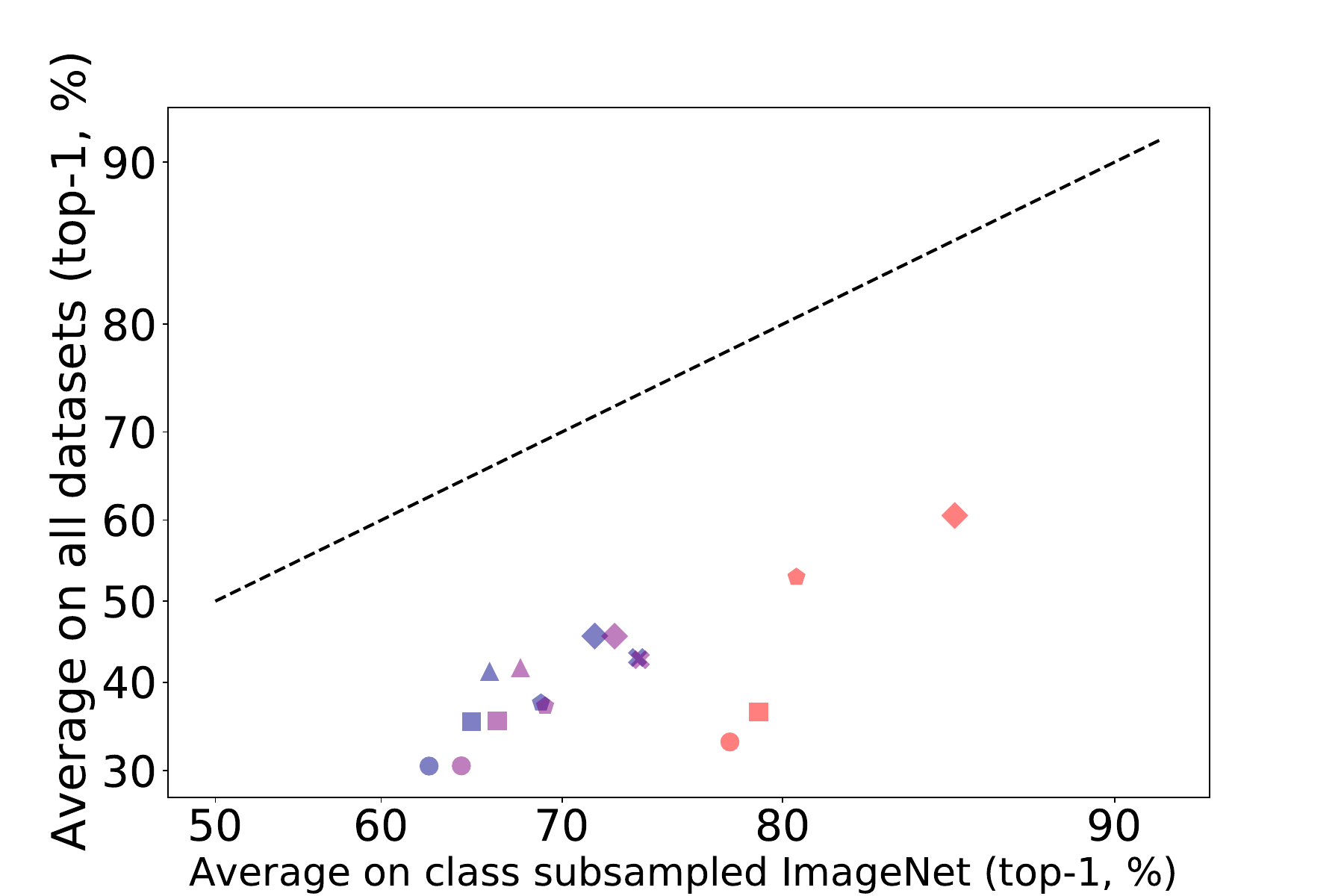}}%
\subcaptionbox{{Two synthetic distribution shifts (ImageNet-C and Stylized ImageNet).\label{fig:alldistb}}}{\includegraphics[width=0.48\textwidth]{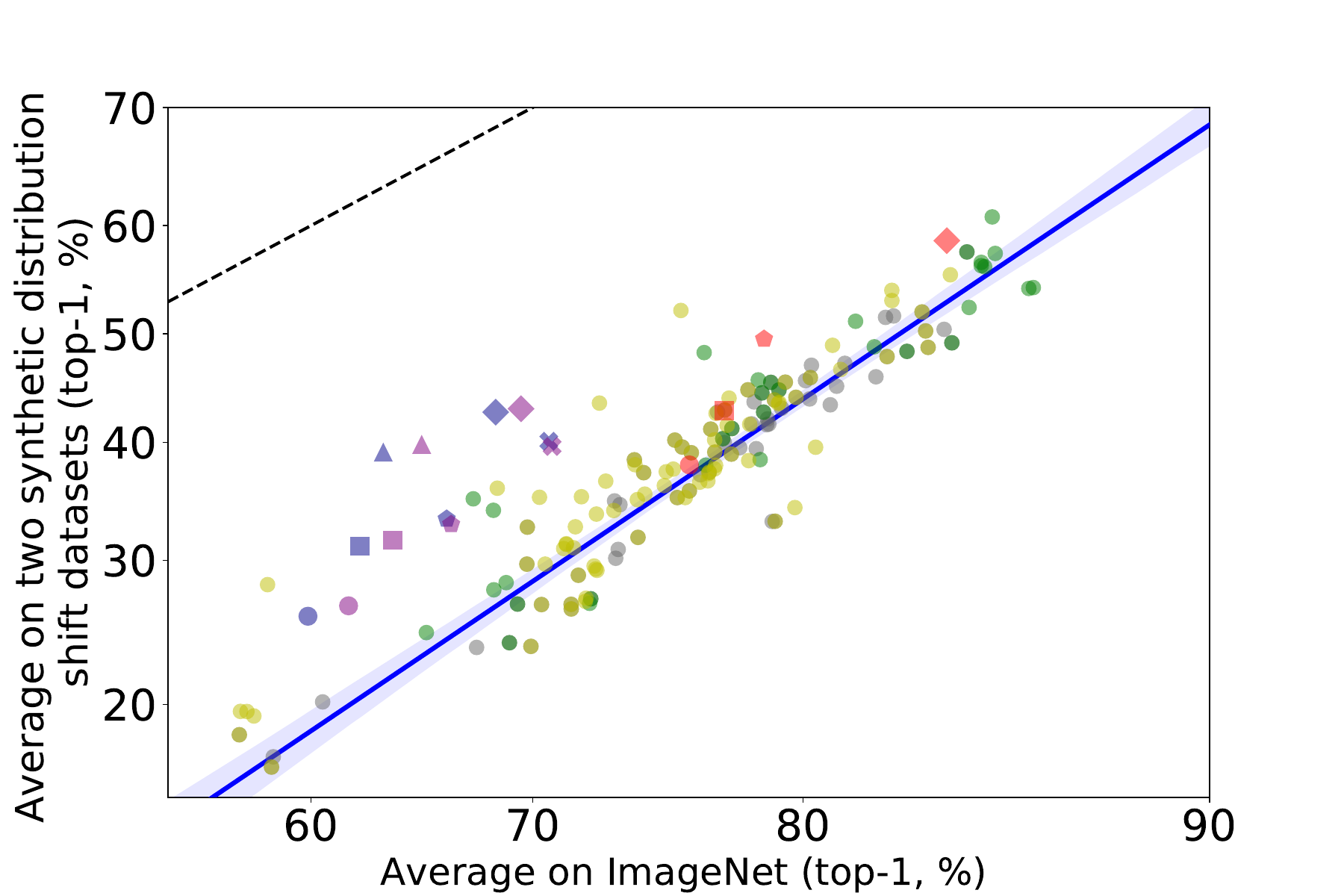}}%
\\
\subcaptionbox{{10 common adversarial attacks on ImageNet and CIFAR-10.\label{fig:alladva}}}{\includegraphics[width=0.48\textwidth]{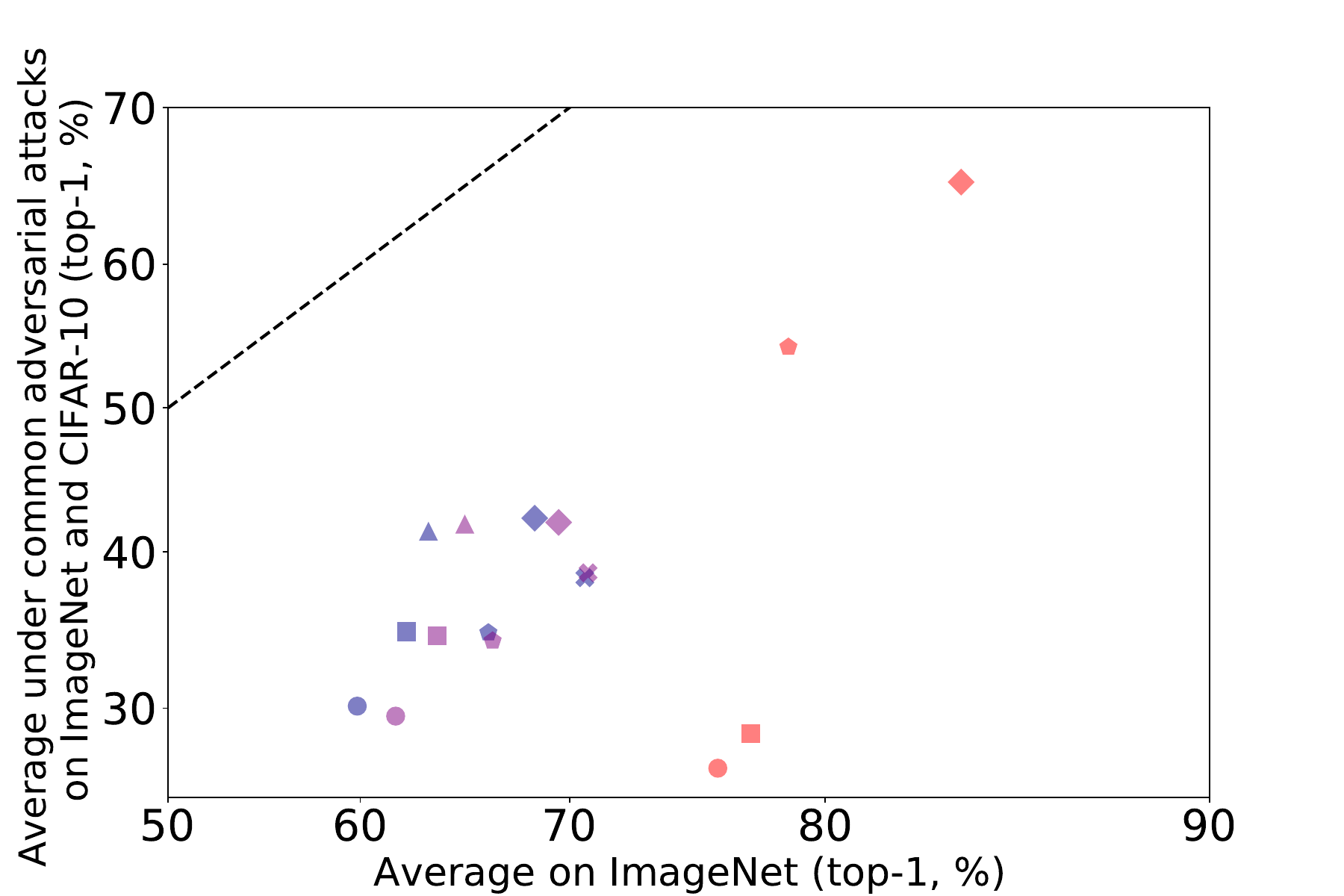}}%
\subcaptionbox{{Typographic attacks on ImageNet-T \ours\ and CIFAR-10-T \ours.}\label{fig:alladvb}}{\includegraphics[width=0.48\textwidth]{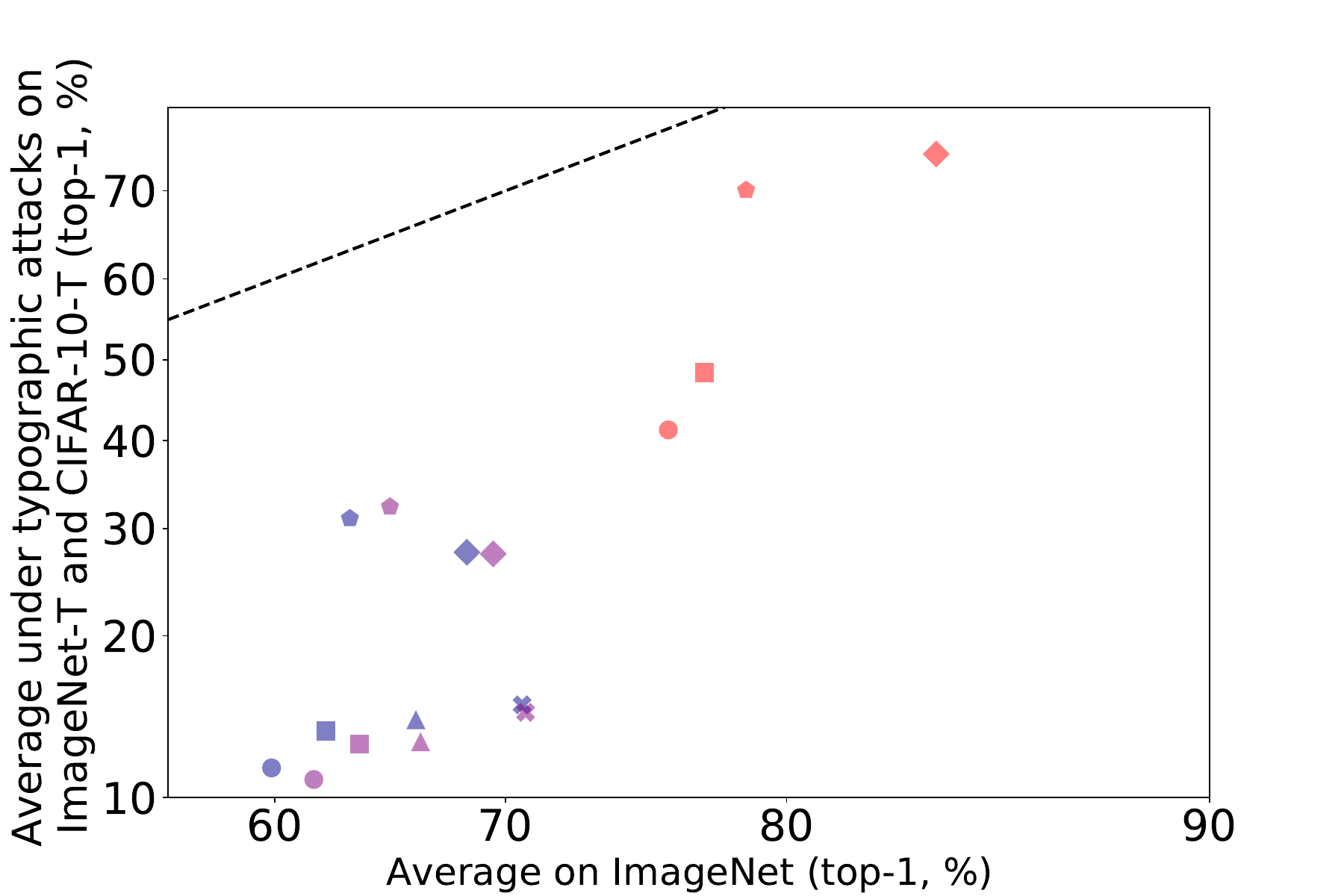}}%
\\
\includegraphics[width=0.95\linewidth]{figs/legend.pdf}%
\vspace{-0.15in}
\caption{{
Zero-shot multimodal CLIP fails to significantly improve the robustness over standard ImageNet models on our \benchmark\ benchmark. 
Red: standard ImageNet models. Blue: zero-shot CLIP models. Purple: CLIP-Auto models. The notable outlier to this trend is CLIP on natural distribution shifts. In particular, we observe a significant performance drop in robustness on our ImageNet-T and CIFAR-10-T. The original CLIP and CLIP-Auto perform similarly on all the test sets.}}
    \label{fig:allres}
\end{figure*}

In this work, we establish a comprehensive robustness benchmark for zero-shot image classification, called \benchmark~(Robustness on Zero-shot), using CLIP as a pilot study. 
Specifically, we make the following contributions. 

\begin{itemize}[leftmargin=*]
\item To systematically evaluate robustness of image classification models, we construct a comprehensive robustness benchmark, \benchmark, that spans 7 natural distribution shifts, 3 synthetic distribution shifts, and 11 adversarial attack models. 

\item Using RoZ, we evaluate the robustness of zero-shot multimodal foundation models, specifically, CLIP. We consider various vision encoders in CLIP, as well as a modified CLIP with automatic prompt generation, as shown in Figure~\ref{fig:introfig}. 

\item While our evaluation of CLIP under natural distribution shift shows robust performance as reported earlier (Figure~\ref{fig:introfig}~(a)), a careful examination of data overlap suggests the observed robustness could be attributed, at least in part, to data overlap.

\item Under synthetic distribution shifts and adversarial attacks,  we show that CLIP leads to significantly downgraded robustness (averaging -11.8\%)  compared to the corresponding standard ImageNet models (Figure~\ref{fig:introfig}~(b)).

\item We introduce a new robustness test set based on the idea of typographic attacks~\cite{goh2021multimodal}, which targets the unique learning paradigm of multimodal learning. CLIP shows a robustness drop of 34.7\%.
\end{itemize}

Our extensive results and analysis suggest that systematic benchmarking in robustness is important to multimodal applications. Our benchmark can be used to evaluate other multimodal models. Furthermore, there is a significant need to improve the robustness of zero-shot multimodal foundation models.

\section{The \benchmark\ Benchmark}

We introduce a benchmark, called \benchmark, to test the robustness of zero-shot multimodal foundation models. The benchmark provides a suite of existing and new robustness datasets. We focus on the zero-shot CLIP model as a pilot study throughout the rest of this work.
\subsection{Datasets and Attacks}
The \benchmark\ benchmark includes common robustness test sets and adversarial attacks. We also create new test sets based on typographic attacks for the benchmark.

\paragraph{Distribution Shifts}
We follow \cite{taori2020measuring} to distinguish two types of distribution shifts. Natural distribution shift refers to the dataset that relies only on natural or unmodified images, while synthetic distribution shift involves modifications of existing images. Image examples of distribution shifts are shown in the appendix.

\begin{itemize}[leftmargin=*]
\item {\bf Natural Distribution Shifts.} We measure on seven natural distribution shifts as in \cite{taori2020measuring}: ImageNetV2~\cite{recht2019imagenet}, ImageNet Sketch~\cite{wang2019learning}, Youtube-BB~\cite{real2017youtube}, ImageNet-Vid~\cite{shankar2019image}, ObjectNet~\cite{barbu2019objectnet}, ImageNet Adversarial~\cite{hendrycks2019natural}, and ImageNet Rendition~\cite{hendrycks2020many}.
\item {\bf Synthetic Distribution Shifts.} We also test on three most widely used synthetic distribution shift datasets: ImageNet-C and ImageNet-P~\cite{hendrycks2019benchmarking}, and Stylized ImageNet~\cite{geirhos2018imagenet}.
\end{itemize}

\paragraph{Adversarial Attacks}
Besides the distribution shifts, we test the robustness to potentially worst-case noises, and adversarial examples.

\begin{figure*}[t]
\centering
\includegraphics[width=0.85\textwidth]{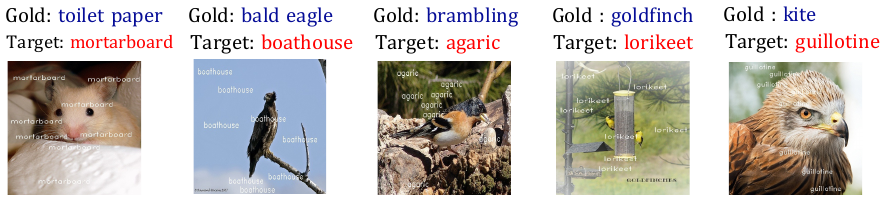}
\vspace{-0.1in}
\caption{{Our ImageNet-T samples. We show the gold class (upper) and the target class (lower) of each sample.}}
    \label{fig:imagenet-T}
\end{figure*}

\begin{itemize}[leftmargin=*]
\item {\bf Common Attacks.} We use 10 widely used image attack methods following \cite{dong2020benchmarking} for the robustness evaluation, including (\expandafter{\romannumeral1}) white-box attacks: FGSM~\cite{goodfellow2014explaining}, DeepFool~\cite{moosavi2016deepfool}, BIM~\cite{kurakin2016adversarial}, and MIM~\cite{dong2018boosting}; (\expandafter{\romannumeral2}) transfer-based attacks: FGSM, BIM, MIM, and DIM~\cite{xie2019improving}; and (\expandafter{\romannumeral3}) black-box attacks: NES~\cite{ilyas2018black} and SPSA~\cite{uesato2018adversarial}. Note that for transfer-based attacks, we use white-box methods on a substitute model to craft adversarial examples. We evaluate the performance on CIFAR-10~\cite{krizhevsky2009learning} and ImageNet~\cite{russakovsky2015imagenet}. 
For CIFAR-10, we utilize the test set of CIFAR-10 containing 10,000 images. For ImageNet, we randomly choose 1,000 images from the ImageNet validation set for evaluation. We focus on untargeted adversarial attacks. We also provide a technical description of the above attack methods in the appendix. 
\item {\bf Typographic Attacks.} Multimodal models are often vulnerable to a kind of non-programmatic adversarial attack, i.e., the typographic attack~\cite{goh2021multimodal}, where adding adversarial text to images can cause them to be systematically misclassified. This is because these models consist of multimodal neurons which respond to both images and texts for a given concept. 
We therefore leverage typographic attacks to specifically examine the robustness of the zero-shot CLIP models. We generate the attacks using the same number of randomly chosen coordinates and using a consistent font style, and focus on targeted adversarial attacks. 
We choose a target class for each image uniformly over all other classes except its true class at random. The target class name is added to each image. For each image in the ImageNet validation set and CIFAR-10 test set, we perform the above image manipulation, resulting in two new robustness datasets: ImageNet Typographic (ImageNet-T) and CIFAR-10 Typographic (CIFAR-10-T). We use 8 and 4 coordinates for the construction of ImageNet-T and CIFAR-10-T, respectively. An example dataset is shown in Figure~\ref{fig:imagenet-T}. 
To the best of our knowledge, this is the first publicly available benchmark based on the concept of typographic attack~\cite{goh2021multimodal}.
\end{itemize}

\subsection{Multimodal Models}
\label{sec:method}
CLIP relies on manual natural language prompts to synthesize the zero-shot image classifier. Besides revisiting zero-shot CLIP, we present CLIP-Auto, which automatically learns prompts for enhanced classification performance.

\paragraph{Zero-Shot CLIP}
CLIP~\cite{radford2021learning} consists of two components: image encoder and text encoder. At a high level, the image encoder is a computer vision backbone that computes a feature representation for the image. The text encoder is a hypernetwork that generates the weights of a linear classifier based on the text specifying the visual concepts that the classes represent. Below is the main architecture of CLIP: (\expandafter{\romannumeral1}) Text encoder architecture: Transformer~\cite{vaswani2017attention} is adopted with the architecture modifications in \cite{radford2019language}. (\expandafter{\romannumeral2}) Image encoder architecture: There are two architectures. The first architecture is ResNet~\cite{he2016deep}. The second architecture is Vision Transformer~\cite{dosovitskiy2020image}. CLIP jointly trains the image encoder and the text encoder to predict the correct pairings of a batch of (image, text) training examples via contrastive learning. At test time, the learned text encoder synthesizes a zero-shot linear classifier by embedding the names or descriptions of the target dataset’s classes. For zero-shot classification on a dataset, CLIP uses the names of all the classes in the dataset as the set of possible text pairs and predicts the most likely (image, text) pairings. CLIP relies on prompt engineering and ensembling to provide manual prompts. Basically, different classifiers are computed based on various manual prompts such as ``A photo of a large \{label\}'' and ``A photo of a little \{label\}''. CLIP ensembles 80 different prompts over the embedding space~\cite{radford2021learning}. We use CLIP in short for the zero-shot CLIP in the remaining paper except noted otherwise.

\paragraph{CLIP-Auto}
While writing prompts is not only time consuming,  it is unclear whether it is optimal for robustness improvements. 
Motivated by the need of prompts that aim to enhance the robustness of zero-shot language models, we adapt the AutoPrompt~\cite{shin2020autoprompt} method to learn better language descriptions for CLIP.
Different from language models that only deal with text, multimodal CLIP handles both images and text for the image classification task. Therefore, our automated prompt combines the image label names with a collection of text trigger tokens, which are learned using a variant of the gradient-based search~\cite{shin2020autoprompt} with respect to image classification loss. The basic idea is that, at each search step, we select a candidate text trigger token to replace a current trigger token that leads to the smallest image classification loss. The classifier computed based on the learned prompt is used to classify the images. We follow \cite{radford2021learning} to ensemble a set of automated prompts to further improve the performance of CLIP. More details are described in the appendix. 

\citet{radford2021learning} have released model settings based on ResNet50, ResNet101, ViT-B/32, ViT-B/16, ResNet50x4, and ResNet50x16. For each setting, we include the corresponding CLIP and CLIP-Auto versions in our evaluations.

\subsection{Comparison Models}
We compare the robustness of the CLIP models to standard models; i.e., image classification models trained on the ILSVRC 2012 dataset~\cite{taori2020measuring}. We focus on the standard models that are parts of the CLIP release for a fair comparison, including: (\expandafter{\romannumeral1}) ResNet~\cite{he2016deep}: ResNet50 and ResNet101; and (\expandafter{\romannumeral2}) Vision Transformer~\cite{dosovitskiy2020image}: ViT-B/32 and ViT-B/16. Note that there is no standard model corresponding to CLIP ResNet50x4 and CLIP ResNet50x16.

We mainly report the results of the above models on our benchmark. We additionally include results of 78 standard models, 86 robust models, and 30 models trained with more data from \cite{taori2020measuring} on distribution shift datasets.

\subsection{Metrics}

We follow \cite{taori2020measuring} to consider two types of robustness: effective and relative robustness.  For a model $m$, we denote two accuracy values: ${\rm acc}_1(m)$ and ${\rm acc}_2(m)$ on a standard test set and a robustness test set, respectively.  Effective and relative robustness are defined as  \cite{taori2020measuring}.

\begin{table*}[t]
\centering
\resizebox{1.0\linewidth}{!}
{
\begin{tabular}{llcccccccccc}
\hline
\multicolumn{2}{l|}{\textbf{Attack Setting}  }                               & \multicolumn{4}{c|}{\textbf{White-Box Attacks}  }                                                                                                                                & \multicolumn{4}{c|}{\textbf{Transfer-Based Attacks}     }                                                                                                                             & \multicolumn{2}{c}{\textbf{Black-Box Attacks}  }                                          \\
\multicolumn{2}{l|}{\textbf{Model} }                                         & \textbf{FGSM}        & \textbf{DeepFool}    & \textbf{BIM}     & \multicolumn{1}{c|}{\textbf{MIM}} & \textbf{FGSM}         & \textbf{BIM}         & \textbf{MIM}          & \multicolumn{1}{c|}{\textbf{DIM}} & \textbf{NES}          & \textbf{SPSA}          \\ \hline
\multirow{3}{*}{ResNet50}   & Standard                              & \best 0.003 / 8.30     & \best  0.0020 / 0.10   & \best 0.002 / 0.10 & \best 0.002 / 0    & \best  0.045 / 54.40    & \best 0.040 / 56.20    & 0.030 / 47.90                          & 0.032 / 49.60                          & \best 0.027 / 57.90     & \best 0.028 / 59.60      \\
                            & CLIP                                  & 0.001 / 5.90                          & 0.0002 / 0                            & 0.001 /  0                        & 0.001 / 0                         &   0.050 / 53.20 &  0.060 / 54.30 & \best 0.047 / 52.30     & \best 0.049 / 51.80     & 0.003 / 30.50                          & 0.003 / 29.80                           \\
                            
                            & CLIP-Auto         &  0.001 / 6.40  & 0.0002 / 0                            & 0.001 / 0                         & 0.001 / 0                         & 0.047 / 52.30                          & 0.057 / 53.40                         &   0.042 / 51.50 &  0.047 / 51.10  &  0.003  / 31.90 &  0.003  /  32.00 \\ \hline
\multirow{3}{*}{ResNet101}  & Standard                              &  0.003 / 8.40  & \best 0.0022 / 0.20    & \best 0.002 / 0    & \best 0.025 / 0    & 0.035 / 52.00                          & 0.033 / 51.70                         & 0.025 / 44.50                          & 0.027 / 46.50                          & \best 0.029 / 57.50     & \best 0.030 / 59.80      \\
                            & CLIP                                  & \best 0.001 / 8.60     & 0.0003 / 0                            & 0.001 / 0                         & 0.001 /  0                        & 0.079 / 55.60                          & 0.078 /  56.70                        &  0.059 / 54.80  & 0.061 / 54.50                          & 0.004 / 36.20                          & 0.004 / 35.80                           \\
                            
                            & CLIP-Auto         & 0.001 / 8.00                          & 0.0003 / 0                            & 0.001 / 0                         & 0.001 /  0                        & \best 0.064 / 56.00     & \best 0.091 / 57.40    & \best 0.062 / 55.10     &  0.067  / 55.50 &   0.005 / 37.20 & 0.005 / 36.60                           \\ \hline
\multirow{3}{*}{ViT-B/32}   & Standard                              & \best 0.006 / 22.00    & \best0.0049 /  9.30    & \best 0.004 / 2.10 & \best 0.004 / 2.00 & \best 0.452 / 76.30     & \best0.748 / 77.00     & \best 0.446 /  76.50    & \best 0.450 / 76.50     & \best0.089 / 72.00      & \best  0.087 / 71.90     \\
                            & CLIP                                  &  0.001 / 10.20 & 0.0008 / 0                            & 0.001 /  0                        & 0.001 / 0                         & 0.117 / 61.00                          & 0.222 / 62.80                         & 0.123 / 61.10                          & 0.139 / 61.50                          & 0.008 /  41.00                         & 0.007 /  40.50                          \\
                            
                            & CLIP-Auto         & 0.001 / 10.10                         &  0.0009 / 0.10 & 0.001 / 0                         & 0.001 / 0                         &  0.111 / 61.80  &  0.243 / 64.00 &  0.143 / 61.50  &  0.145 / 62.60  &  0.010 /  43.70 &  0.010 / 43.10   \\ \hline
                \multirow{3}{*}{ViT-B/16}   & Standard                              & \best0.005 / 16.10 &
\best0.004 / 3.70 &
\best0.004 / 0.70 &
\best0.004 / 0.50 &
\best0.474 / 78.90 &
\best0.800 / 80.20 &
\best0.471 / 79.30 &
\best0.452 / 79.20  & 
\best0.095 / 75.50 &
\best0.098 / 75.60  \\ 
                            & CLIP                                  & 0.001 / 6.70 &
0.0009 / 0 &
0.001 / 0&
0.001 / 0&
0.169 / 62.70&
0.221 / 64.60&
0.149 / 62.60&
0.152 / 62.70&
0.011 / 42.80&
0.013 / 43.50  \\
                           
                            & CLIP-Auto         & 0.002 / 6.30&
0.0010 / 0&
0.001 / 0&
0.001 / 0&
0.171 / 64.10&
0.2265 / 65.10&
0.151 / 64.40&
0.1535 / 64.40&
0.011 / 40.30&
0.012 / 39.80               \\ \hline
\multirow{2}{*}{ResNet50x4} & CLIP                                  & \best 0.001 / 12.20    & 0.0003 / 0                            & 0.001 / 0                         & 0.001 /  0                        & \best 0.115 / 57.50     & \best 0.122 / 60.40    & \best 0.080 / 56.50     & \best 0.085 / 57.20     & 0.006 / 37.60                          & 0.006 / 37.30                           \\
                            & CLIP-Auto         &  0.001 / 11.70 & \best 0.0004 / 0.10    & 0.001 / 0                         & 0.001 /  0                        &  0.079 / 56.90  &  0.119 / 58.80 &  0.074 / 56.10  &  0.078 / 56.60  & \best 0.006 / 40.70     & \best  0.007 / 40.60     \\ \hline
                    
\multirow{2}{*}{ResNet50x16} & CLIP                                  &  0.001 / 14.60&
0.0005 / 0&
0.001 / 0&
0.001 / 0&
0.216  / 63.40&
0.188 / 64.60&
0.113 / 61.80&
0.122 / 62.40&
\best 0.014 / 49.00&
\best 0.014 / 48.40 \\
                           
                            & CLIP-Auto         &\best   0.001 / 15.10&
0.0005 / 0&
0.001 / 0&
0.001 / 0&
 \best0.213 / 64.20&
 \best0.173 /  66.40&
 \best0.108 /  63.30&
 \best0.125 / 63.60&
0.012 /  48.70&
0.013 /  48.20     \\ \hline

\end{tabular}
}
\caption{{Model results against individual untargeted adversarial attacks under the $l_{\infty}$ norm on ImageNet. Each entry consists of the median $l_{\infty}$ distance of the minimum adversarial perturbations over all samples on the left, and the model accuracy for the perturbation budget $\epsilon = 8/255$ on the right. Note that there are no corresponding standard ResNet50x4 and ResNet50x16 available. We highlight the results based on accuracy.}}
\label{tab:imgNetadv}

\end{table*}

\paragraph{Effective Robustness} 
Instead of directly comparing accuracy, effective robustness aims to measure how much higher accuracy on the robustness test sets is compared to the accuracy on the standard test set. Formally, the effective robustness of a model is defined as: ${\rm acc}_2(m) - \beta({\rm acc}_1(m))$, where $\beta(\cdot)$ is the baseline accuracy on a robustness test set for a given accuracy on the standard set. Graphically, effective robustness corresponds to a model being above the trend (blue line) given by a set of standard ImageNet models in Figure~\ref{fig:alldista}. $\beta(\cdot)$ indicates the blue line.

\paragraph{Relative Robustness} 
Effective robustness does not measure the improvements brought by a robustness technique. Therefore, relative robustness directly measures the improvements on the robustness test sets. Formally, given a model $m$ and its robustness enhanced version $m'$, the relative robustness is ${\rm acc}_2(m') - {\rm acc}_2(m)$.

A robust model should be able to improve both effective and relative robustness. To help analyze the effective and relative robustness, we report the average accuracy across the corresponding datasets and average accuracy across the corresponding class subsets of ImageNet (Figure~\ref{fig:introfig} and Figure~\ref{fig:allres}). We use the average of pm-0 and pm-10 accuracy for Youtube-BB and ImageNet-Vid. We average over the five severities for each corruption in ImageNet-C. We use mFR and mT5D on ImageNet-P. We also report the median $l_{\infty}$ distance of the minimum adversarial perturbations across all samples for the adversarial attacks, and the success rate on ImageNet-T and CIFAR-10-T.

\section{Results}
\label{sec:resultsection}

In this section, we show that our \benchmark\ benchmark provides a comprehensive robustness evaluation of multimodal CLIP models. Except for natural distribution shifts, CLIP generally fails to improve robustness over the corresponding standard models on our benchmark. We first summarize the main results (Sec.~\ref{sec:mainres}), then describe the breakdown results with a focus on synthetic distribution shifts (Sec.~\ref{sec:syndist}) and adversarial attacks (Sec.~\ref{sec:adv}). Details about the additional experimental setups and results are described in the appendix.

\subsection{Main Results} 

\label{sec:mainres}
CLIP fails to improve the robustness over the corresponding standard models in image classification. In Figure~\ref{fig:all}, we compare the average accuracies of the zero-shot CLIP models with their CLIP-Auto versions and standard ImageNet models on all datasets. We find that most robustness improvements of CLIP are due to the significant improvements on natural distribution shift, in particular the effective robustness. The result on natural distrubiton shift is similar to that reported in \cite{radford2021learning}. Note that one vision model is evaluated for robustness in \cite{radford2021learning} while we evaluate multiple vision models and show that they all demonstrate similar behaviors. In Figure~\ref{fig:alldista}, we summarize the performance of zero-shot CLIP models compared to existing ImageNet models and CLIP-Auto models on natural distribution shifts. The details are shown in Table~\ref{table:natural}. All CLIP models improve the effective robustness over standard ImageNet models on natural distribution shifts. These CLIP models also improve the relative robustness of the standard models on the natural distribution shifts except for one dataset, ImageNetV2. ImageNetV2 follows the original creation process of ImageNet, which suggests that the distribution is likely to be similar to the ImageNet distribution, and thus the standard models in general work well.

However, we draw contrary conclusions on the rest of our benchmark: synthetic distribution shifts and adversarial examples. CLIP has lower average robustness on these test sets. In particular, CLIP models are much more vulnerable to typographic attacks than standard models, resulting in a substantial 34.74\% performance drop on average. We find that CLIP-Auto does not make much difference in the robustness performance compared to CLIP. This is in contrast to the conclusion in pre-trained language models (e.g., T5~\cite{raffel2019exploring} and GPT-3~\cite{brown2020language}). The reason is that  the image representation learned from large-scale pre-trained data is the key differentiator of the classification performance. The optimization of prompts that synthesize the linear classifiers on top of the image representation has limited impact. We find Vision Transformer is at least as robust as CLIP. We also find that while the effective robustness is comparable, CLIP actually reduces the relative robustness by a considerable amount compared to the corresponding standard model.

\begin{table*}[t]
\centering
\resizebox{0.9\linewidth}{!}{
\begin{tabular}{llcccccccc}
\hline
\multicolumn{2}{l}{\textbf{Model}}               & \textbf{ImageNet} & \textbf{ImageNetV2} & \textbf{ImageNet-R} & \textbf{ObjectNet} & \textbf{ImageNet-Sketch} & \textbf{ImageNet-A} & \textbf{Youtube-BB} & \textbf{ImageNet-Vid} \\ \hline
\multirow{3}{*}{ResNet50}   & Standard           & \best 76.13                              & \best 62.70                                & 35.05                                & 35.77                               & 22.20                                 & 0.81                                 & 50.10                                & 60.14                                  \\
                            & CLIP               & 59.85                              & 52.64                                & \best 60.51                                & \best 39.82                               &  35.46                                &  22.75                                & \best 56.69                                & \best 64.92                                  \\
                            
                            & CLIP-Auto  &  61.78                              & 54.53                                &  60.48                                & 39.21                               & \best35.48                                & \best 23.81                                & 56.68                                & 64.61                                  \\ \hline
\multirow{3}{*}{ResNet101}  & Standard           & \best 76.20                           & \best 64.30                               & 37.70                                & 32.60                               & 25.20                                & 2.70                                 & 53.10                                & 62.80                                  \\
                            & CLIP               & 62.32                              & 56.08                                & \best 68.01                                & 44.17                               & \best 41.09                                &  29.44                                &  61.75                                &  71.28                                  \\
                            
                            & CLIP-Auto  &  63.83                              &  56.95                                & 67.37                                & \best 45.01                               &  40.89                                & \best 30.36                                & \best 62.11                                & \best 71.64                                  \\ \hline
\multirow{3}{*}{ViT-B/32}   & Standard           &  \best 78.73                              & \best 71.16                                &  44.69                                & \best 43.76                               & 32.55                                &\best 33.64                                & \best 63.62                                & \best 77.72                                  \\
                            & CLIP               & 63.40                              & 55.73                                & \best 69.29                                &  43.60                               & \best 42.42                                & 31.35                                & 60.98                                & 74.03                                  \\
                            
                            & CLIP-Auto  &  65.16                              &  57.51                                &  68.10                                & 43.16                               &  42.15                                &  32.21                                &  61.01                                &  74.80                                  \\ \hline
\multirow{3}{*}{ViT-B/16}   & Standard           &  \best 84.20                              & \best 74.13 &	50.89 &	51.12	& 38.10 &	\best 50.64 &	\best 64.76 & \best 81.79                                   \\
                            & CLIP               & 66.94                           & 62.52 & \best	77.82 &	\best 53.45 &	\best 48.47 &	49.39 &	 64.07    & 80.26                        \\
                            
                            & CLIP-Auto  &  69.51                             &  63.11	&  76.83	 &  53.08 & 	 48.43 &	 49.56	& 63.28	 & 79.76                         \\ \hline
\multirow{2}{*}{ResNet50x4} & CLIP               &  66.28                              & 59.34                                & \best 72.63                                &  49.97                               & \best 44.75                                & \best 41.53                                &  59.53                                & \best 72.50                                  \\
                            
                            & CLIP-Auto  & \best 66.48                              & \best 59.75                                & 71.33                                & \best 50.36                               &  44.55                                &  40.95                                & \best 59.75                                & 72.14                                  \\ \hline

\multirow{2}{*}{ResNet50x16} & CLIP               &  70.67                              & \best  64.14 &	\best  79.18 &	 58.86	& \best 50.66 & 	\best 56.41 & 	\best 62.96 &	\best 78.45 \\
                            
                            & CLIP-Auto  & \best 70.81                            &  63.90 &	 78.33 &	\best  59.04 &	 49.62 &	 55.93 &	 61.97 &	 78.04                               \\ \hline
\end{tabular}%
}
\vspace{-0.1in}
\caption{{Model accuracy on ImageNet and seven natural distribution shifts.}}
        \label{table:natural}
\end{table*}

\begin{figure*}
\centering
\subcaptionbox{{ImageNet-C.}\label{fig:imagenetCbreakdown}}{\includegraphics[width=0.48\textwidth]{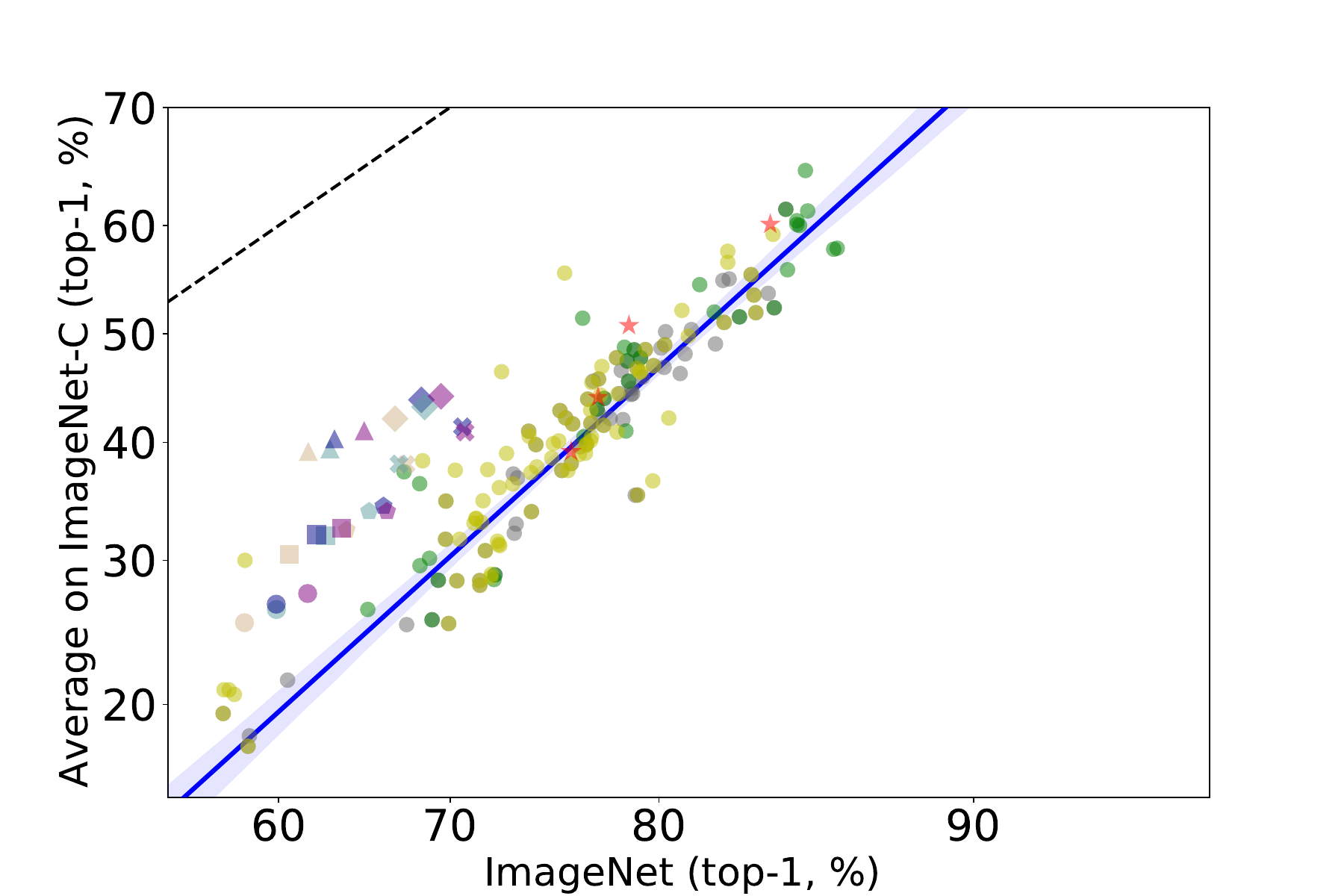}}
\hspace{0.02in}
\subcaptionbox{{Stylized ImageNet.}\label{fig:stylizedimagenetbreakdown}}{\includegraphics[width=0.48\textwidth]{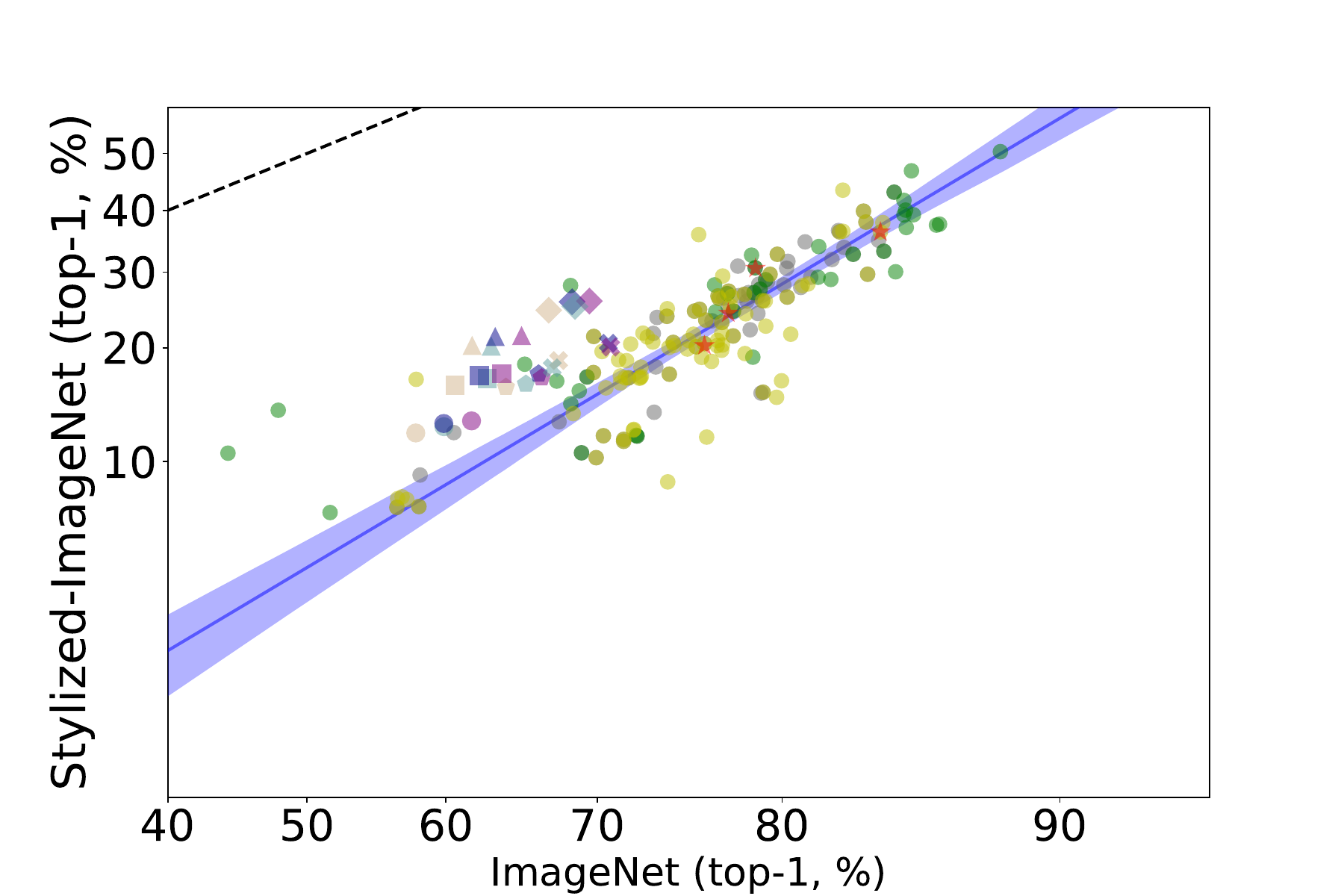}}%
\\
\includegraphics[width=0.95\linewidth]{figs/legend.pdf}%
\vspace{-0.2in}
\caption{{Model accuracies on two synthetic distribution shifts. Different from the results on natural distribution shifts, we show that CLIP fails to improve the robustness compared to standard models. Red:  standard ImageNet models. Blue: zero-shot CLIP models. Purple: CLIP-Auto models.}}
\label{fig:syntheticbreakdown}
\end{figure*}

\subsection{Synthetic Distribution Shifts}
\label{sec:syndist}

In Figure~\ref{fig:alldistb} and Figure~\ref{fig:syntheticbreakdown}, we summarize the model accuracies on two synthetic distribution shifts, ImageNet-C and Stylized ImageNet. 
We show detailed results on ImageNet-C and ImageNet-P in the appendix. 
We observe downgraded robustness of CLIP compared to the corresponding standard models. 
We also find that the performance of CLIP and CLIP-Auto are comparable, again suggesting that prompt enhancement does not improve its robustness. The reason is that the key ingredient to the robustness of multimodal CLIP is pre-trained image representations, while the impact of language prompts is limited. This is different from the conclusion in language models, as language prompts improve robustness in a single modal setup. Also, we suspect that this is due to images of synthetic distribution shifts are not in the pre-training data and present a further analysis in Sec.~\ref{sec:dataoverlap}. 
Improving the zero-shot robustness to synthetic distribution shifts via regularization techniques is one of our future investigations. 

\subsection{Adversarial Attacks}
\label{sec:adv}

\begin{table}
\centering

\resizebox{0.95\linewidth}{!}{
\begin{tabular}{llcc}
\toprule
\multicolumn{2}{l}{\multirow{2}{*}{\textbf{Model}}}             & \textbf{CIFAR-10-T}                         & \textbf{ImageNet-T}                       \\
\multicolumn{2}{l}{}                                            & \multicolumn{1}{l}{\textbf{Success rate / Accuracy}} & \textbf{Success rate / Accuracy} \\ \midrule
\multirow{3}{*}{ResNet50}   & Standard         & \best  8.16 / 13.90                      & \best  0.03 / 68.66                       \\
                            & CLIP             & 99.40 / 0.22                      & 40.42 / 22.61                      \\
                                             
                            & CLIP-Auto        & 99.46 / 0.18                      & 44.86 / 21.50                      \\ \midrule
\multirow{3}{*}{ResNet101}  & Standard         & \best 6.49 / 27.03                      & \best 0.03 / 69.70                       \\
                            & CLIP             &  97.97 / 1.61                      &  37.76 / 25.25                      \\
                                 
                            & CLIP-Auto         & 98.07 / 1.45                      & 42.47 / 23.93                      \\ \midrule

\multirow{3}{*}{ViT-B/32}   & Standard         & \best 3.73 / 65.12                      &\best  0.02 / 75.03                       \\
                            & CLIP             & 77.47 / 17.08                     & 13.07 / 45.06                      \\
                                           
                            & CLIP-Auto        & 75.24 / 17.67                     &  12.64 / 47.01                      \\ \midrule
\multirow{3}{*}{ViT-B/16}   & Standard         & \best 2.10 / 67.42                   &\best  0.02 / 80.05                      \\
                            & CLIP             & 89.84 / 9.51                     & 25.07 / 45.59                      \\
                                              
                            & CLIP-Auto        & 90.54 / 8.79                    &  25.14 / 45.99                     \\ \midrule

\multirow{2}{*}{ResNet50x4} & CLIP             & 98.75 / 1.06                      & \best 44.21 / 27.14                      \\
                            
                            & CLIP-Auto        & \best 99.13 / 0.72                      & 48.28 / 24.90                      \\ \midrule

\multirow{2}{*}{ResNet50x16} & CLIP             & 97.85 / 2.07                     & \best 50.22 / 27.99                     \\
                           
                            & CLIP-Auto        & \best  97.59 / 2.30                     &   52.29 / 26.80                   \\ \bottomrule
\end{tabular}}
\vspace{-0.1in}
    \caption{{Model accuracies and success rates under typographic attacks on ImageNet-T \ and CIFAR-10-T. We highlight the results based on accuracy. CLIP causes a significant robustness drop compared to the corresponding standard models.}}
    \label{tab:typo}
  
\vspace{-0.1in}
\end{table}
\paragraph{Common Attacks}
We test the robustness to 
adversarial examples, which is crucial for safety-critical applications. 
We compare the average accuracies under common adversarial attacks in Figure~\ref{fig:alladva}. 
On CIFAR-10, we use zero-shot CLIP, CLIP-Auto, and linear probe standard models for the attacks. 
The CLIP models are more vulnerable when compared to the standard models. We illustrate results on ImageNet in Table~\ref{tab:imgNetadv}. The trends on ImageNet and CIFAR-10 are similar. 
CLIP-Auto does not improve the adversarial robustness over CLIP. The reason is that the pre-training stage does not include adversarial examples and is not robust to adversarial attacks. A further discussion is presented in Sec.~\ref{sec:dataoverlap}. Improving the zero-shot robustness to adversarial attacks via adversarial prompt learning is an important future direction. Additional results are shown in the appendix.

\begin{figure*}
\centering
\subcaptionbox{{ ImageNet-T.}\label{fig:typographicimgNet}}{\includegraphics[width=0.45\textwidth]{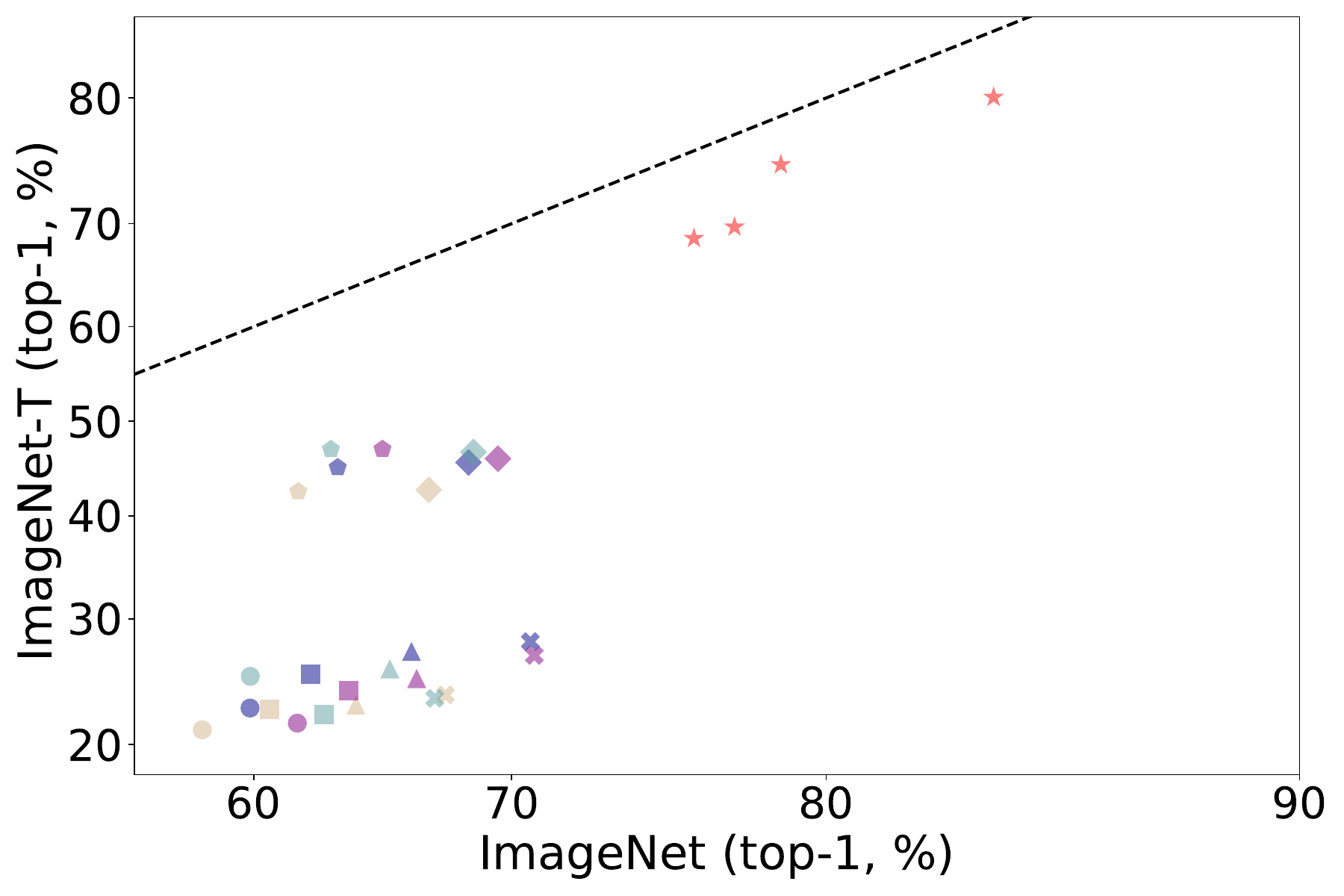}}%
\hspace{0.02in}
\subcaptionbox{{ CIFAR-10-T.}\label{fig:typographiccifar}}{\includegraphics[width=0.45\textwidth]{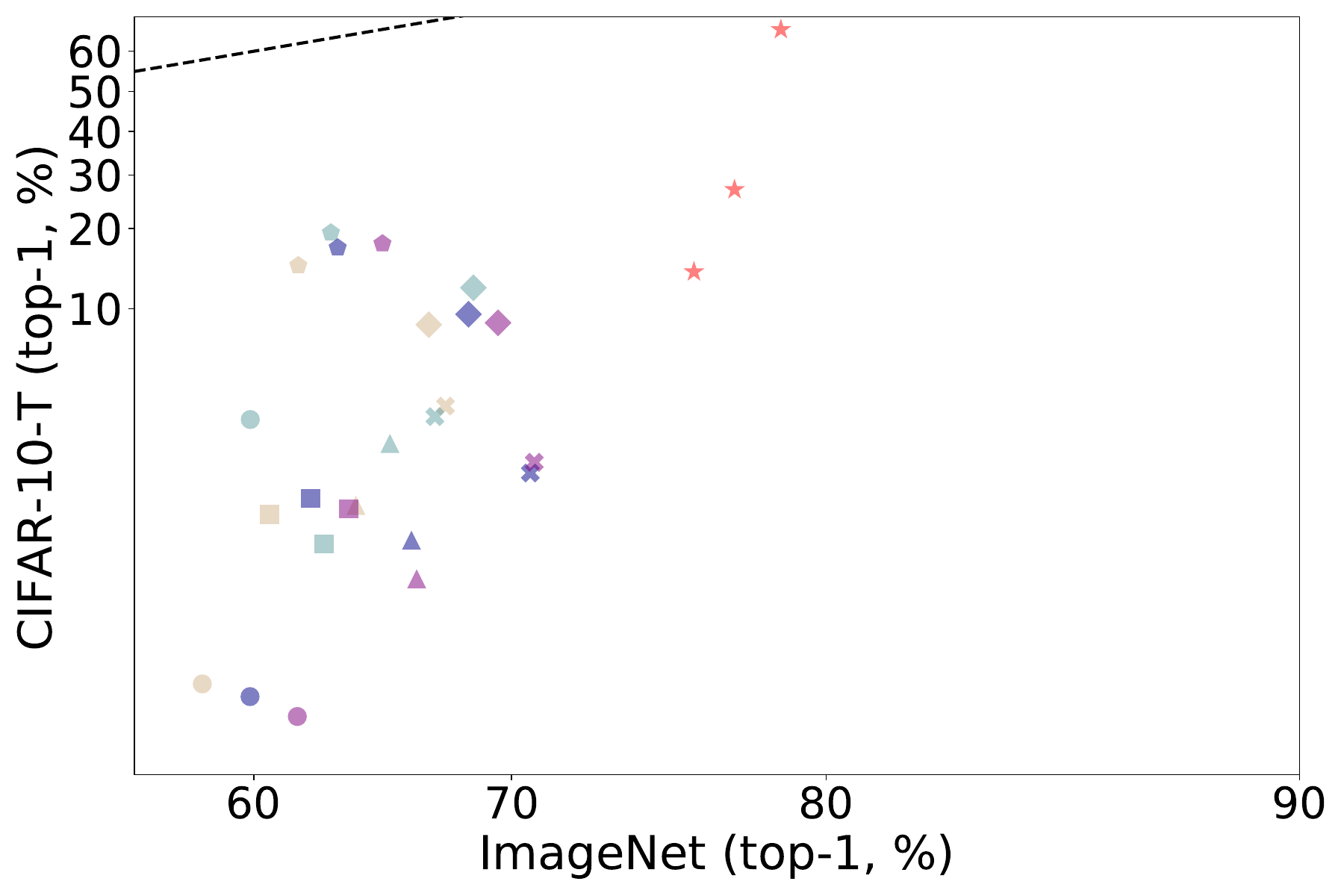}}

\includegraphics[width=0.95\linewidth]{figs/legend.pdf}%
\vspace{-0.2in}
\caption{{ Model accuracies under typographic attacks on our ImageNet-T and CIFAR-10-T. Red: standard ImageNet models. Blue: zero-shot CLIP models. Purple: CLIP-Auto models.}}

\label{fig:typobreakdown}
\end{figure*}

\paragraph{Typographic Attacks}
CLIP is extremely vulnerable to our new robustness test sets (ImageNet-T and CIFAR-10-T) that are based on a new kind of non-programmatic attack named typographic attacks \cite{goh2021multimodal}.
In Figure~\ref{fig:alladvb} and Figure~\ref{fig:typobreakdown}, we find that CLIP reduces both effective and relative robustness by a large amount (-34.74\% in average accuracy) compared to the standard models on ImageNet-T and CIFAR-10-T. The attack success rate is also much higher for the CLIP models (Table~\ref{tab:typo}). The underlying reason is that, different from standard models, multimodal CLIP learns to respond to both images and text given a concept. Adding adversarial text to images can fool the CLIP models. This also applies to CLIP-Auto, as learnable prompts still correspond to a visual concept. We plan to improve the zero-shot robustness to typographic attacks via regularization techniques to force CLIP to only focus on image representation. 
The results indicate ImageNet-T and CIFAR-10-T are important test sets for understanding zero-shot robustness.

\section{Data Overlap Analysis}
\label{sec:dataoverlap}
As shown in Sec.~\ref{sec:resultsection}, while CLIP achieves improved robustness on natural distribution shifts, it fails to transfer to other robustness test sets in \benchmark\ benchmark. A growing problem when training high-capacity models on large-scale datasets is data contamination, where the pre-training dataset can potentially include content from the test datasets because such content is on the web. We suspect that the data contamination issue in the pre-training data actually results in the performance on natural distribution shifts.

Although CLIP~\cite{radford2021learning} has conducted data overlap analysis, we find the analysis is not rigorous since it assumes the overlapped images share the same distribution with the test sets. We propose to rigorously measure the data overlap between the CLIP pre-training data and the robustness test sets. The main idea is to remove image examples that are the same or similar to training examples from test sets. The ``cleaned'' test sets can be used for robustness re-evaluation. In particular, we use the image encoder of ResNet50x16 as the duplication detector as it is trained on the same distribution as the pre-training set. The deduplication threshold is defined as the cosine similarity between image representations. We consider the images where the similarity between them are beyond the deduplication threshold as overlapped images, which are then removed from the test sets. Since the entire pre-training set of CLIP has not been released, we use a subset of it, YFCC100M dataset~\cite{thomee2016yfcc100m}. We focus on the comparison between natural and synthetic distribution shifts.

\begin{figure*}
\centering
\subcaptionbox{{Percentage of overlapped images on natural or synthetic distribution shifts.}\label{fig:sample_ratio}}{\includegraphics[width=0.45\textwidth]{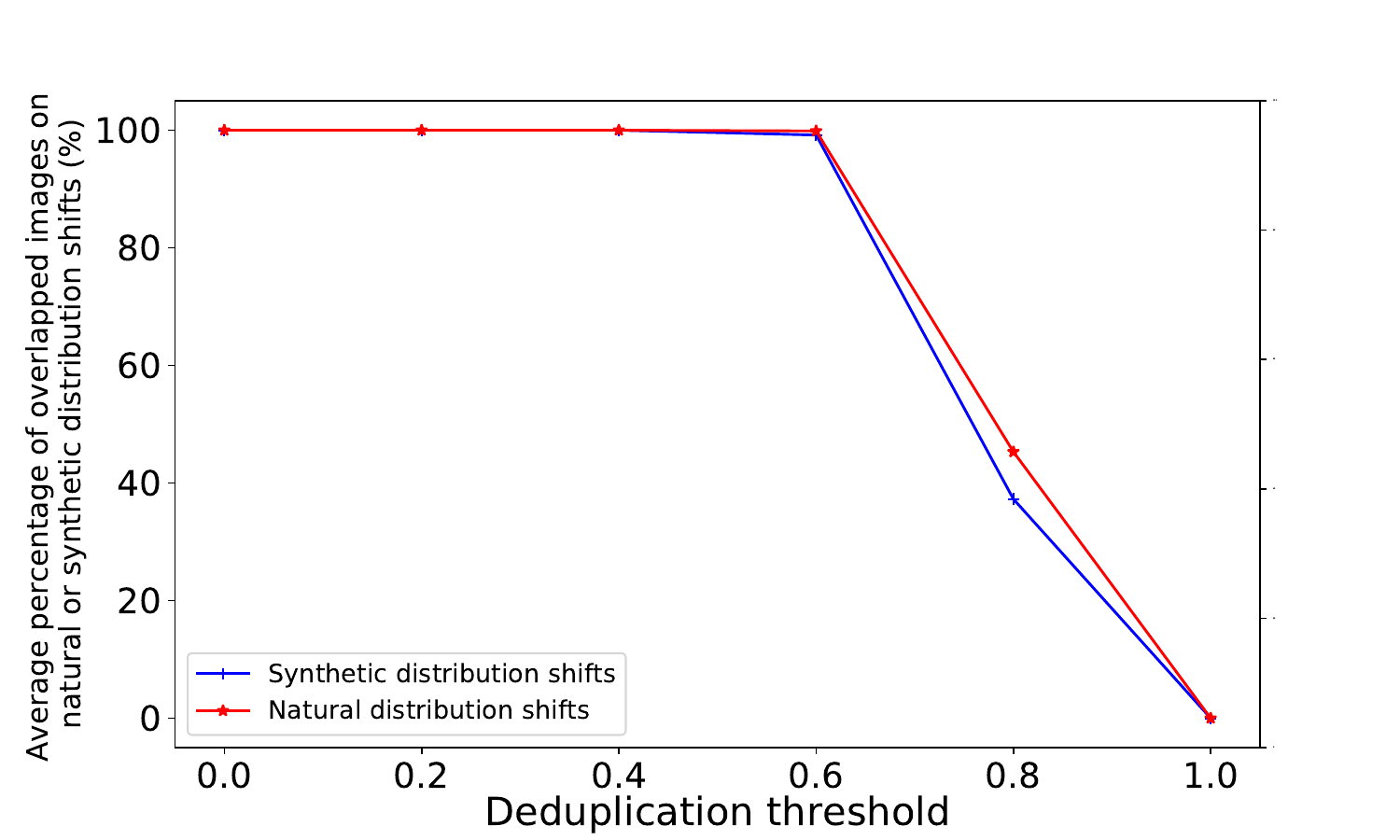}}
\hspace{0.06in}
\subcaptionbox{{Percentage of overlapped images and corresponding accuracies on ImageNetV2 and Stylized ImageNet. Left y-axis: Accuracy. Right y-axis: Percentage. }\label{fig:deduplicationb}}{\includegraphics[width=0.45\textwidth]{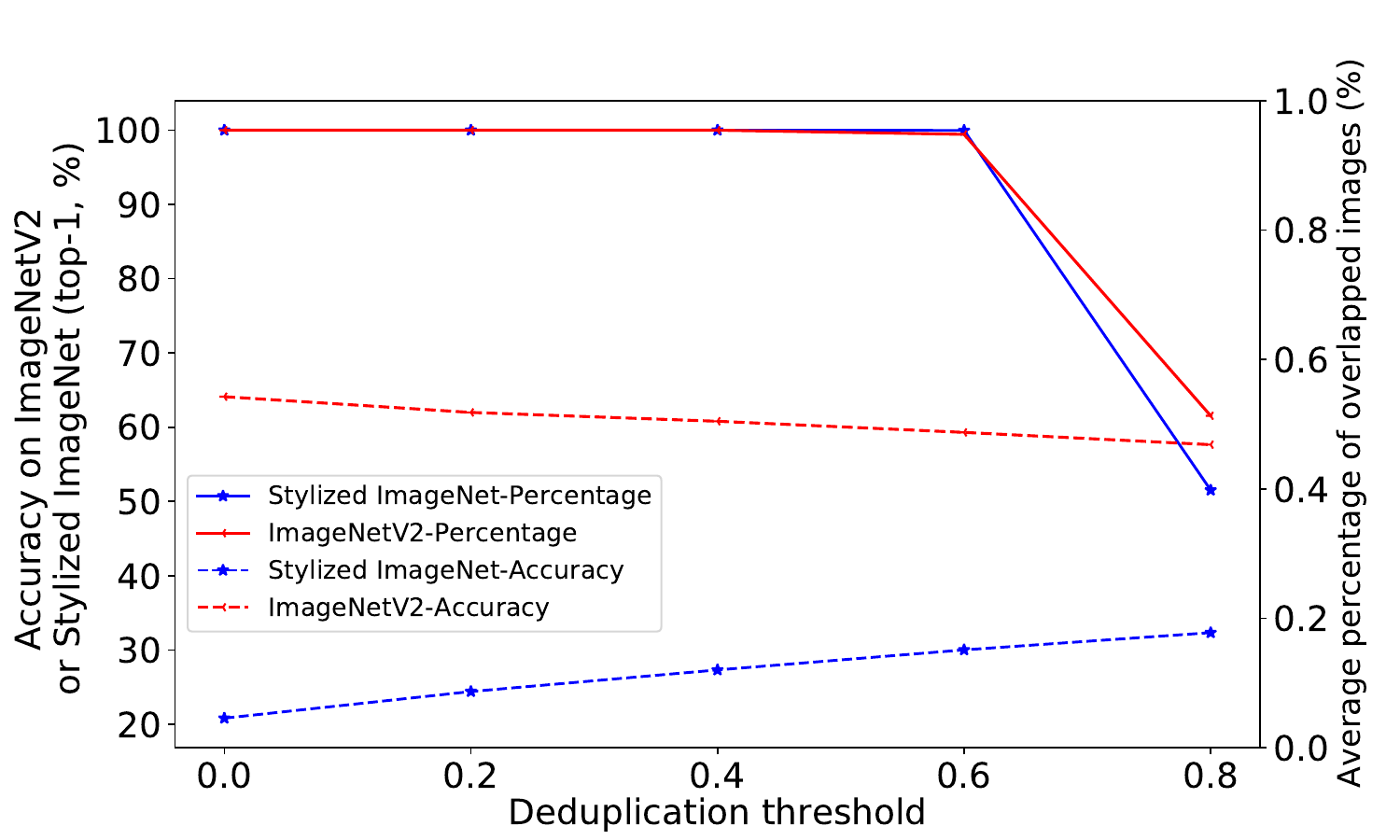}}%
\\
\vspace{-0.1in}
\caption{{Data overlap between YFCC100M (pre-training data) and distribution shifts.}}
\label{fig:deduplication}
\end{figure*}

Figure~\ref{fig:sample_ratio} and Figure~\ref{fig:deduplicationb} show the overall comparison and a case study. Importantly, while performance drops on ImageNetV2, we see an accuracy improvement on Stylized ImageNet. This indicates that the natural distribution shift benefits from the data overlap as the pre-training set contains similar images. While the above analysis is based on a small subset of original CLIP training data, more severe data overlap issues may exist when testing on the full training data. Besides, we simulate the pre-training data using Google Images, and find that very similar (or even the same) images are found by Google Images through querying with an image in ImageNetV2. The results are presented in the appendix. We argue that it is crucial to clean the pre-training data to test robustness, which is a common practice when training large-scale models. For example, GPT-3~\cite{brown2020language} has reported significantly inflated results due to the data overlap issue. Our probe is an initial attempt to understand the role of data overlap on robustness. We hope that the analysis promotes further research along this line.

\section{Related Work}
\citet{radford2019language} focus on the robustness of CLIP to natural distribution shifts. \citet{taori2020measuring} conduct a comprehensive robustness study of the image classification models on both natural distribution shifts and synthetic distribution shifts. \citet{dong2020benchmarking} evaluate the adversarial robustness of image classification models. \citet{koh2021wilds} propose a new natural distribution shifts datasets. \citet{shen2024measuring} propose a new vision-language STEM understanding dataset. Different from these studies, we focus on conducting a comprehensive study of the zero-shot robustness in the domain of image classification, considering all natural distribution shifts, synthetic distribution shifts, and worst-case adversarial examples. In addition, we benchmark the robustness under a new kind of attack, typographic attacks. To the best of our knowledge, this is the first work to quantitatively measure the robustness under typographic attacks.

\citet{radford2021learning} test the zero-shot robustness using the default natural language prompts~\cite{crispino2023agent}. We measure the robustness by disentangling the effect of the natural language prompts in the evaluation. Zero-shot robustness of language models has found pre-training improves the relative robustness~\cite{hendrycks2020pretrained}, but little evidence of effective robustness improvements~\cite{miller2020effect}. In comparison, we test the zero-shot robustness in image classification. 

\citet{li2017learning} use the learned visual n-grams to perform zero-shot image classification. VirTex~\cite{desai2020virtex}, ICMLM~\cite{sariyildiz2020learning} and ConVIRT~\cite{zhang2020contrastive} have demonstrated the usage of natural language in learning image representations. \citet{gomez2017self} and \citet{joulin2016learning} also introduce methods that learn visual representations from natural language supervision. In this work, we focus on the current state-of-the-art model learned from the text, CLIP. There are similar models including GLIP~\cite{GLIP}, GLIDE~\cite{GLIDE}, and BLIP~\cite{li2023blip}. We leave the study of the robustness of other work as a future direction.

\section{Conclusion}
We construct a comprehensive benchmark to study the zero-shot robustness of multimodal foundation models using CLIP as a pilot study. Our results show that CLIP is not robust under synthetic distribution shifts and adversarial attacks, and its previously reported robustness under natural distribution shifts might be attributed, at least in part, to data overlap. The finding differs from the original finding in the CLIP paper, where they conclude that the model is more robust than standard models trained on ImageNet. 
In order to benefit real-world applications, especially in the safety-critical fields, our results suggest that it is crucial to conduct comprehensive robustness evaluations. 
We hope our results will foster further research into the zero-shot robustness of foundation models.

\section*{Acknowledgements}
We would like to thank Yangyi Chen for the early inputs.

{
    \small
    \bibliographystyle{ieeenat_fullname}
    \bibliography{main}
}

\clearpage
\appendix

\section{The RoZ Benchmark Details}

\paragraph{Distribution Shifts Settings}
\begin{itemize}[leftmargin=*]
    \item {\bf Natural Distribution Shifts}. We follow the settings in \cite{taori2020measuring}. For ImageNetV2, we report results on imagenetv2-matched-frequency-format-val. For ImageNet-A, ImageNet-R, Youtube-BB, ImageNet-Vid, and ObjectNet, we only take predictions that are also in the category of the ImageNet validation set. We use the same setup for ImageNet Sketch as in \cite{taori2020measuring}.
    \item {\bf Synthetic Distribution Shifts}. For ImageNet-C, we use all 15 common corruption types including: gaussian noise (on disk), shot noise (on disk), impulse noise (on disk), defocus blur (on disk), glass blur (on disk), motion blur (on disk), zoom blur (on disk), snow (on disk), frost (on disk), fog (on disk), brightness (on disk), contrast (on disk), elastic transform (on disk), pixelate (on disk), jpeg compression (on disk). For each corruption, we average over the five severities. For ImageNet-P, we use 10 common perturbations: gaussian noise, shot noise, motion blur, zoom blur, snow, brightness, translate, rotate, tilt, and scale. We also use the Stylized ImageNet dataset~\cite{geirhos2018imagenet}.
\end{itemize}

\paragraph{Common Adversarial Attack Methods} Based on the different levels of knowledge of the target model, we consider the following attack scenarios from white-box attacks that have access to the model architectures and parameters, to transfer-based attacks and black-box attacks that only have access to the training data or model outputs. There are two typical strategies for creating adversarial examples with small perturbations. The first results in adversarial examples with a constrained perturbation, while the second strategy produces an adversarial example with an optimized perturbation.
\begin{itemize}[leftmargin=*]
    \item {\bf White-Box Attacks}. White-box attacks rely on detailed information of the target model. White-box attacks craft adversarial examples based on the gradient of the input. We include the following widely-used attack methods: fast gradient sign method (FGSM)~\cite{goodfellow2014explaining}, basic iterative method (BIM)~\cite{kurakin2016adversarial}, DeepFool~\cite{moosavi2016deepfool}, and momentum iterative method (MIM)~\cite{dong2018boosting}. We empirically set the number of iterations to 12 and 20 for both BIM and MIM in two strategies respectively. We set the maximum number of iterations as 50 for DeepFool in two strategies.
    \item {\bf Transfer-Based Attacks}. The attacks have access to the training data and leverage the adversarial transferability~\cite{papernot2016practical}, aiming to obtain a substitute model from which the adversarial examples are created. We craft adversarial examples from the above white-box methods on a substitute model including FGSM, BIM, and MIM. Besides, we incorporate the diverse inputs method (DIM)~\cite{xie2019improving} to improve adversarial transferability. We use a ResNet-152 model~\cite{he2016deep} as the substitute model following \cite{dong2020benchmarking}. We set the number of iterations to 12 and 10 for all methods in both strategies. 
    \item {\bf Black-Box Attacks}. Black-box attacks, in particular, score-based black-box attacks only have access to output probabilities via querying the target model. Therefore the gradient can be estimated by gradient-free methods. We include NES~\cite{ilyas2018black} and SPSA~\cite{uesato2018adversarial} that conduct gradient estimation based on random samples and the corresponding loss. We set the number of iterations to 12 and 20 for both methods in two evaluation settings.
\end{itemize}
To ease the reproductivity, we use the same hyperparameters as \cite{dong2020benchmarking} for all methods. We refer readers to \cite{akhtar2018threat} for a survey of the attack methods.

\paragraph{Implementation Details} For all distribution shift datasets, we leverage the image classification testbed~\cite{taori2020measuring}\footnote{\small{\url{https://modestyachts.github.io/imagenet-testbed/}}} to evaluate the results. The testbed includes all the standard models. We integrate the released CLIP models\footnote{\small{\url{https://github.com/openai/CLIP}}} and the corresponding CLIP-Auto models into the testbed. For adversarial attack experiments, we implement all adversarial attack methods for the evaluation. For all experiments, we use: 4 GeForce GTX 1080 with a batch size of 32 per GPU. The average runtime is approximately 60 minutes on seven natural distribution shifts, 10 minutes on ImageNet-P and Stylized-ImageNet, 450 minutes on ImageNet-C, and 45 minutes under 11 adversarial attacks.

\section{More Results}

\begin{figure*}
\centering
\subcaptionbox{{ImageNetV2.}\label{fig:imagenetV2}}{\includegraphics[width=0.31\textwidth]{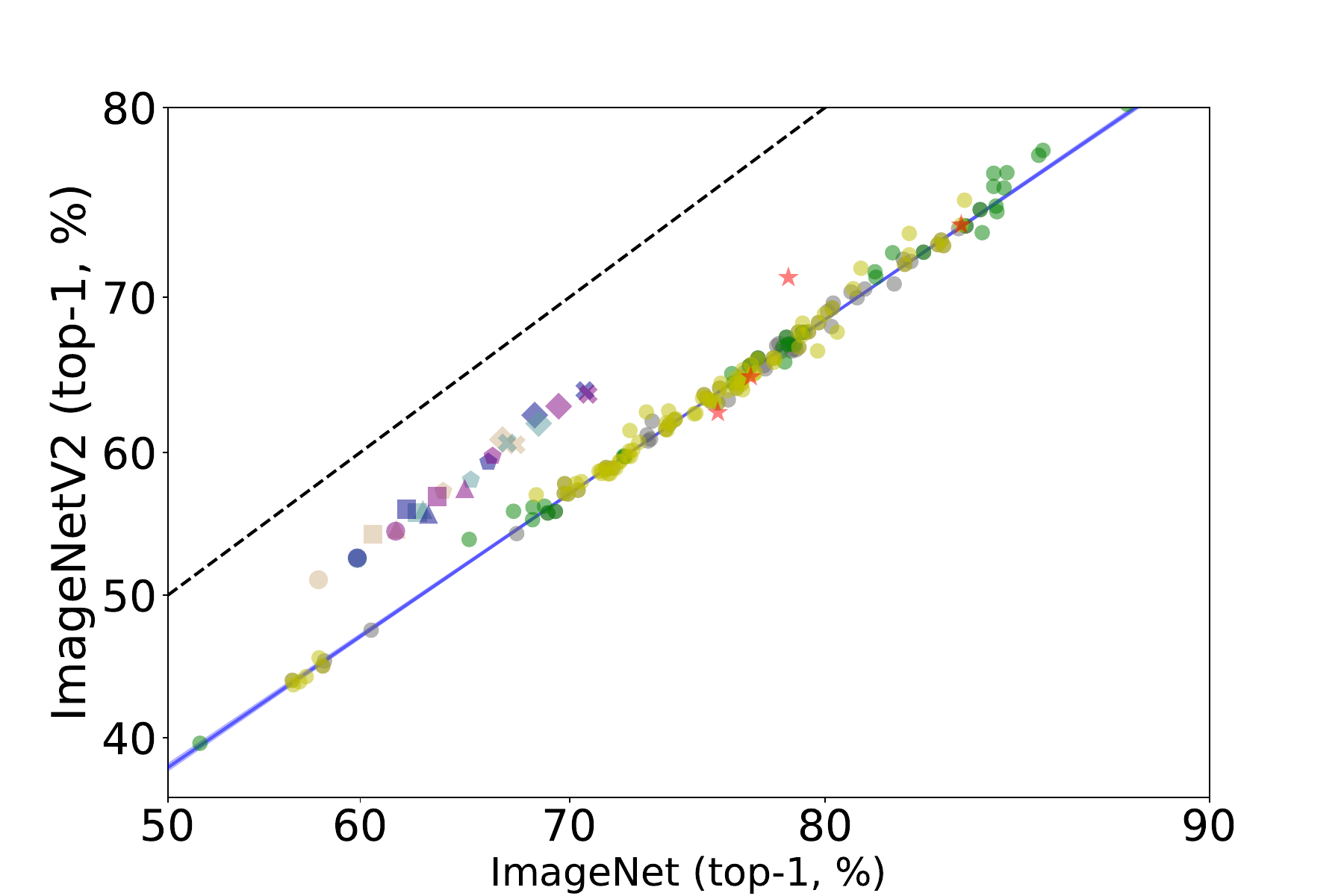}}%
\hspace{0.02in}
\subcaptionbox{{ImageNet-R.}\label{fig:imagenet-R}}{\includegraphics[width=0.31\textwidth]{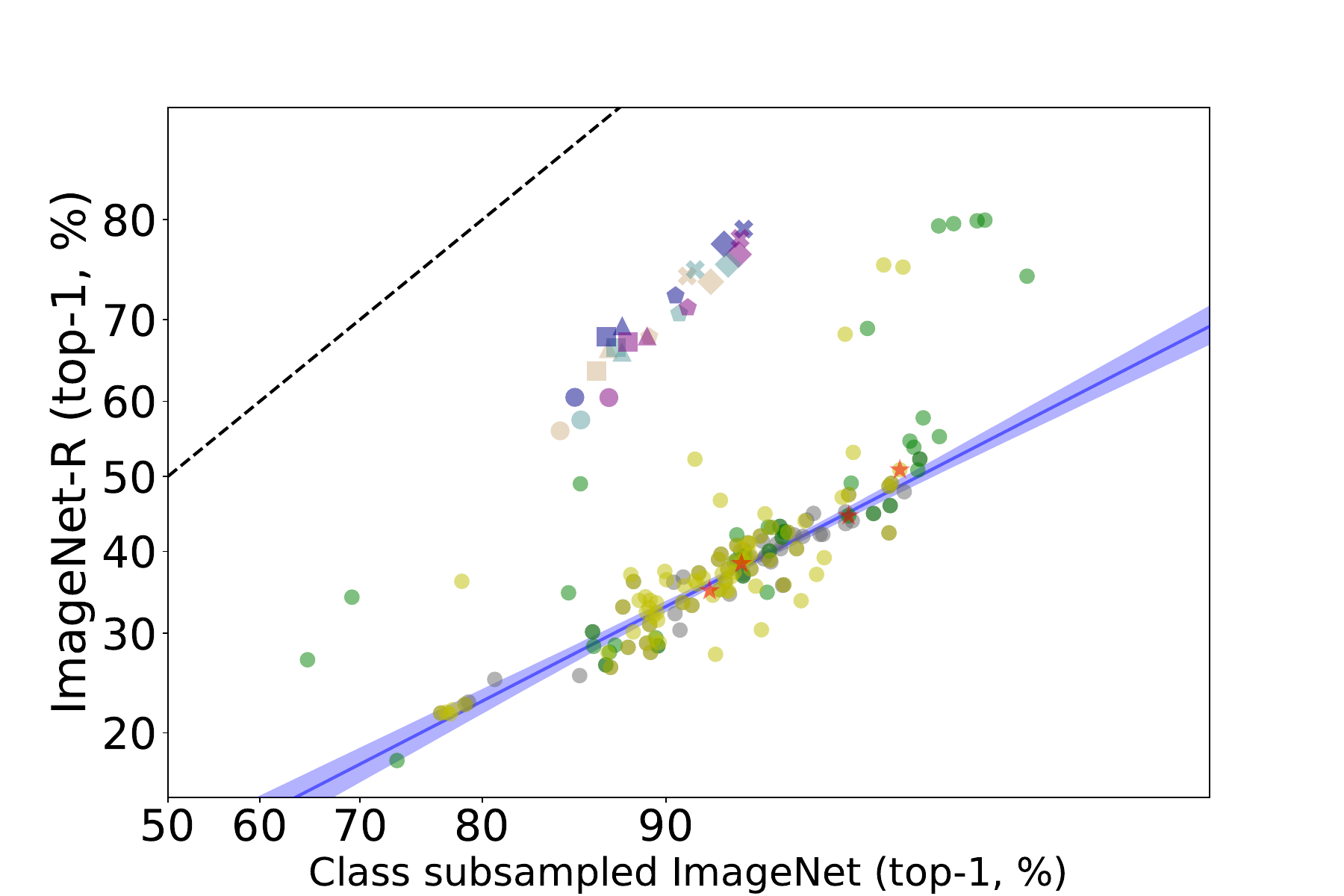}}%
\\

\subcaptionbox{{ObjectNet.}\label{fig:objectnet}}{\includegraphics[width=0.31\textwidth]{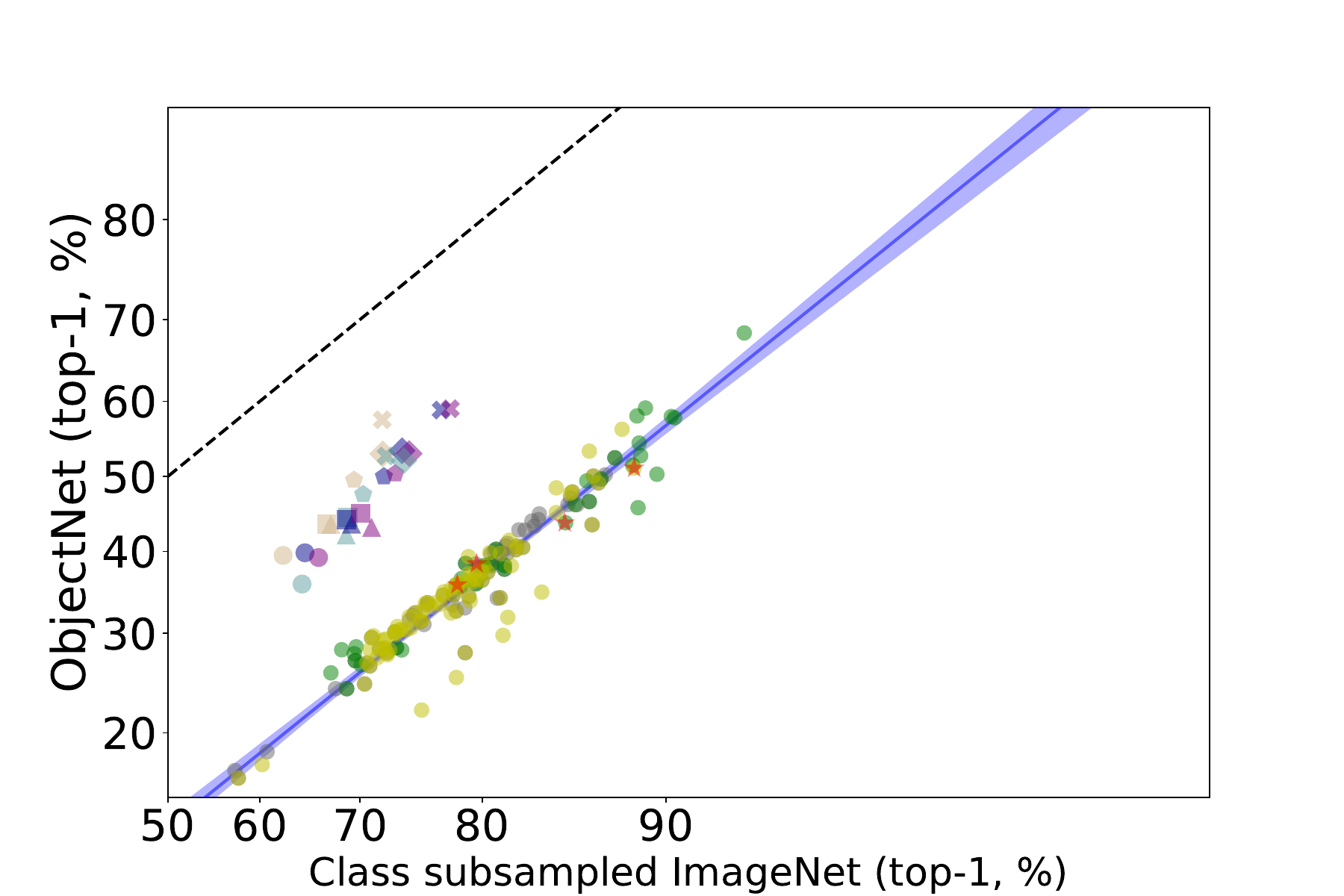}}%
\hspace{0.02in}
\subcaptionbox{{ ImageNet-Sketch.}\label{fig:imagenet-sketch}}{\includegraphics[width=0.31\textwidth]{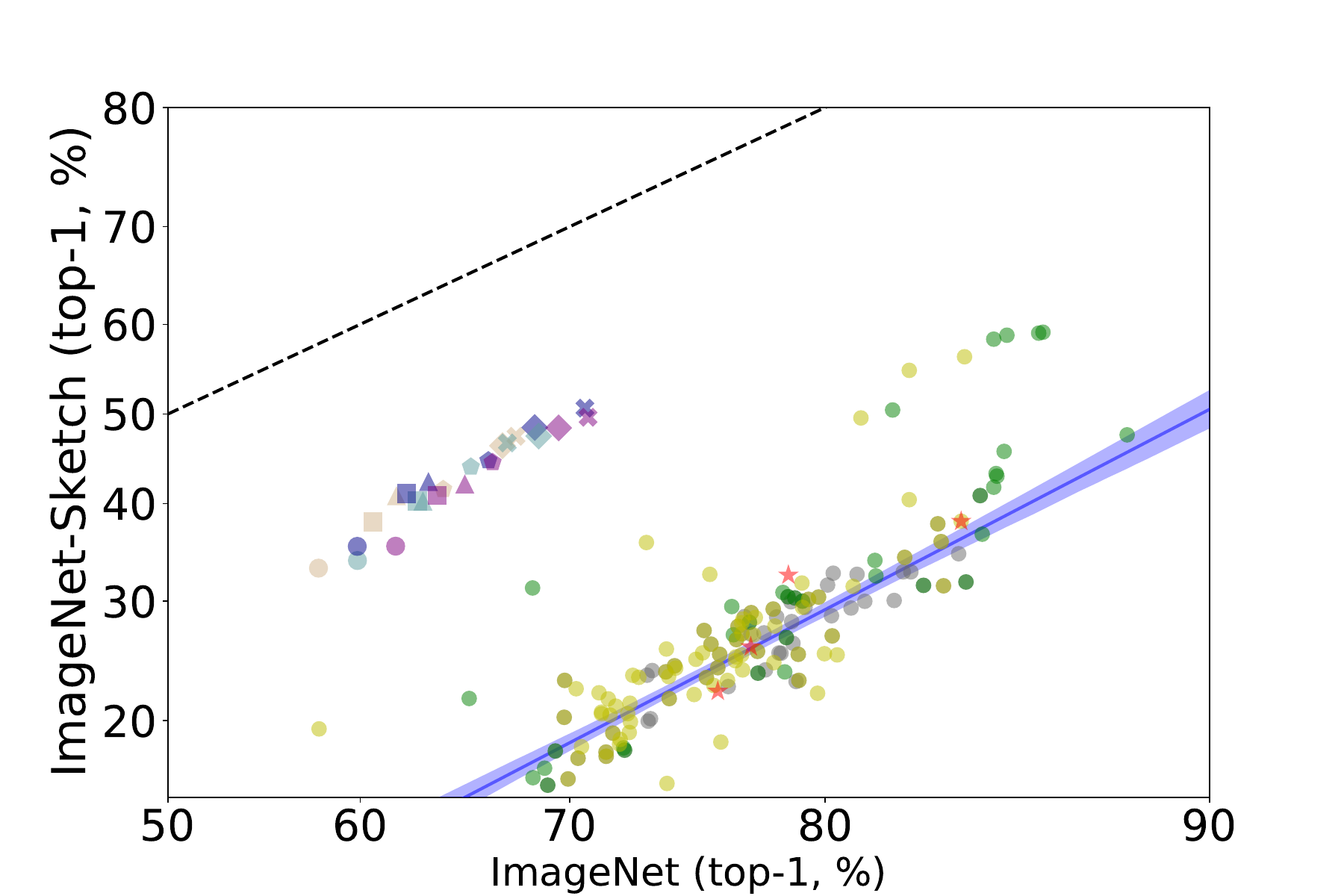}}%
\\

\subcaptionbox{{ ImageNet-A.\label{fig:imagenet-a}}}{\includegraphics[width=0.31\textwidth]{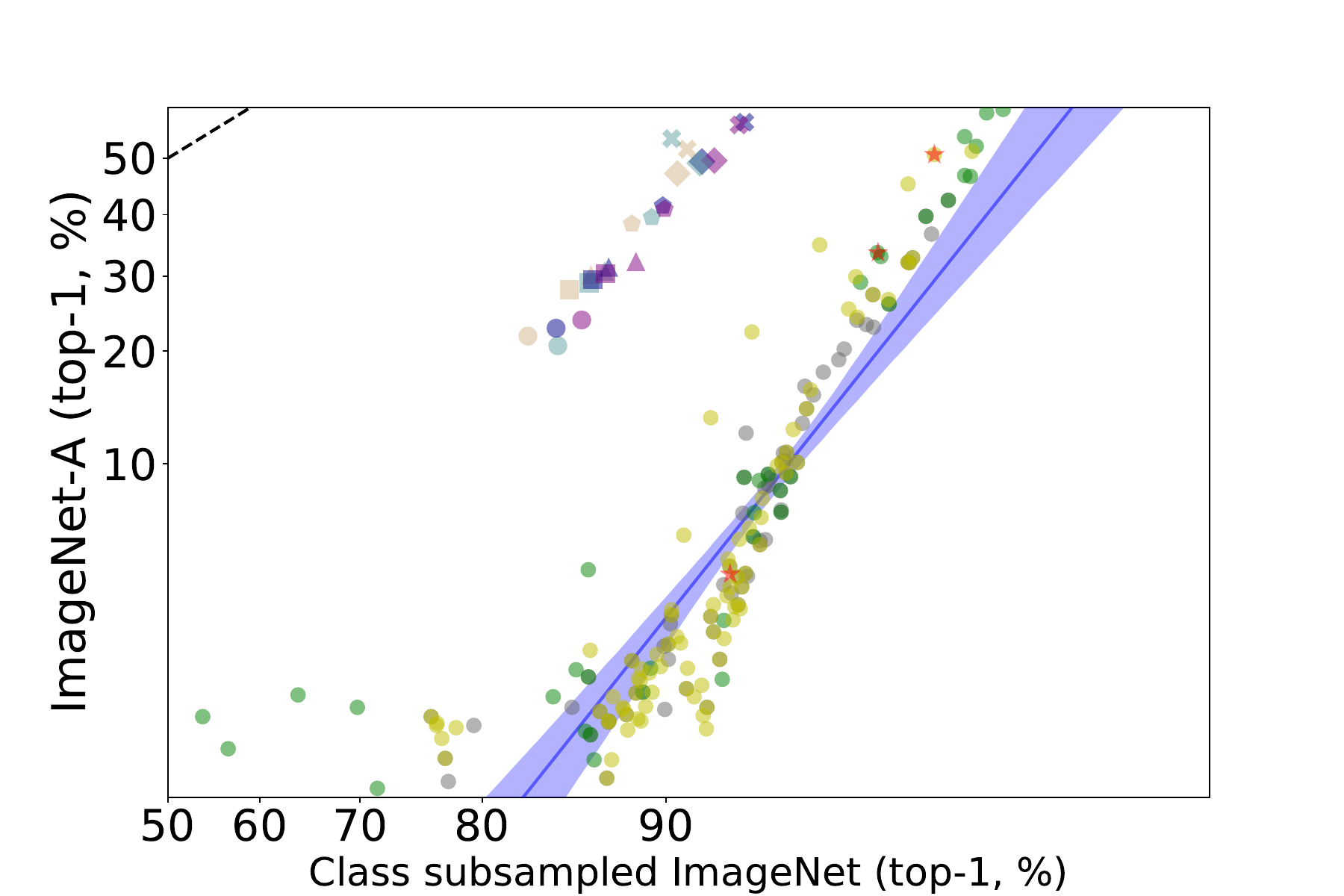}}%
\hspace{0.02in}
\subcaptionbox{{ Youtube-BB. \label{fig:imagenet-ytbb}}}{\includegraphics[width=0.31\textwidth]{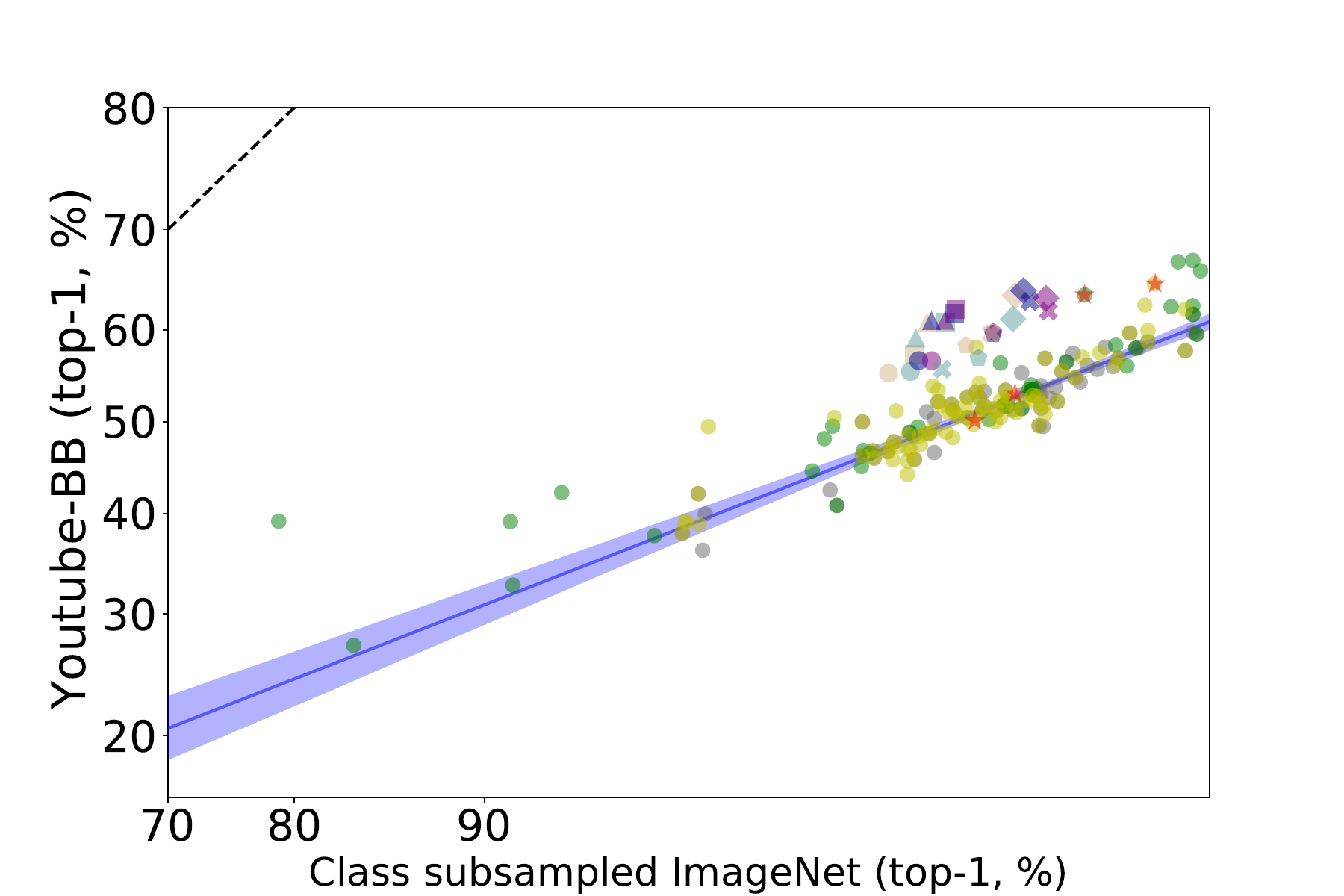}}%
\hspace{0.02in}
\subcaptionbox{{ ImageNet-Vid.}\label{fig:imagenet-vid}}{\includegraphics[width=0.31\textwidth]{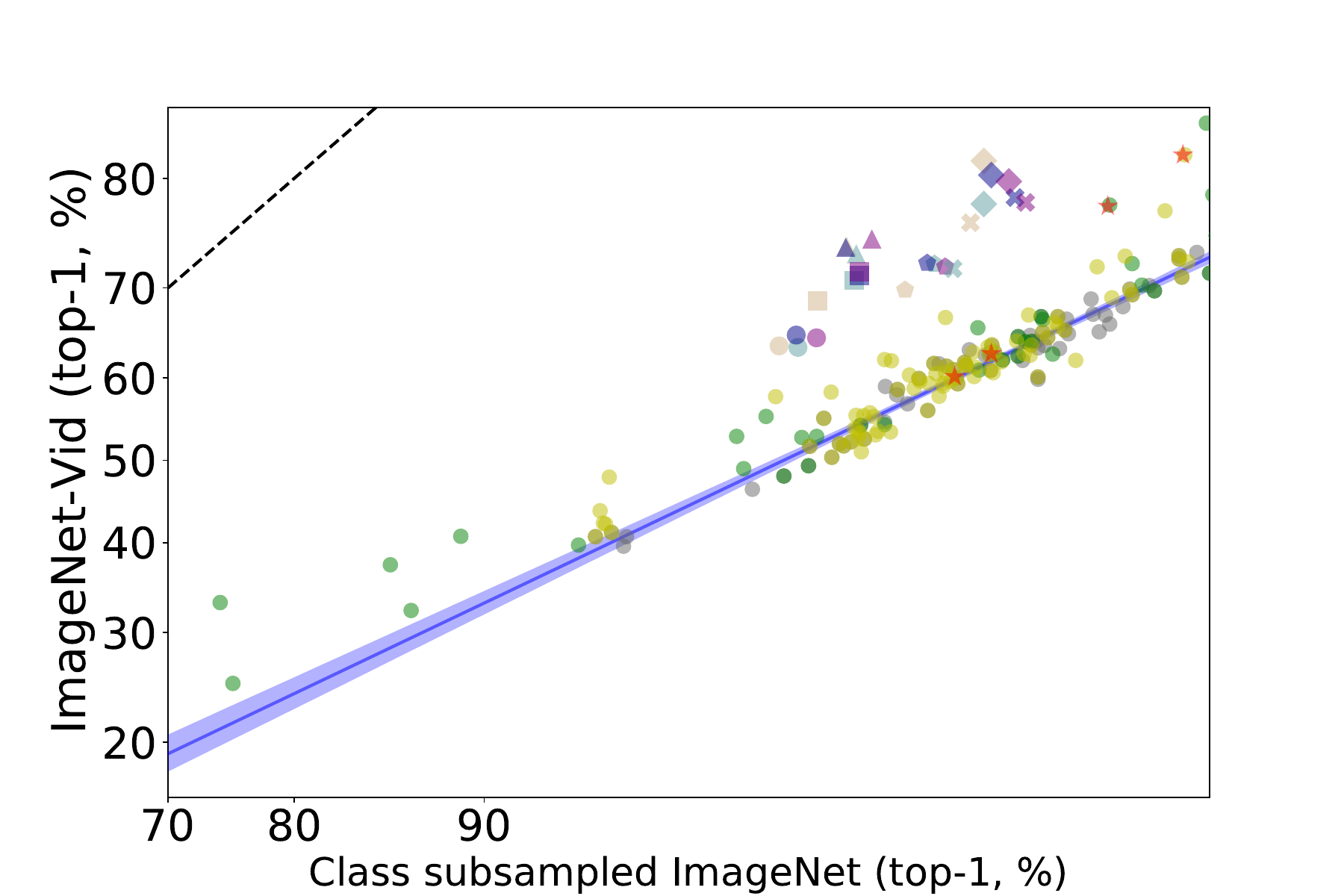}}%
\\
\includegraphics[width=0.95\linewidth]{figs/legend.pdf}%
\caption{{ Model accuracies on the seven natural distribution shifts. Red:  standard ImageNet models.  Blue:  zero-shot CLIP models.  Purple: CLIP-Auto models.}}
    \label{fig:ndistbreakdown}
\end{figure*}

Sec.~\ref{sec:resultsection} provides a high-level summary of the robustness results, we show the breakdown results on each individual dataset in our \benchmark \   benchmark in this section.

\begin{table}[]
\centering
\resizebox{0.5\linewidth}{!}{
\begin{tabular}{llc}
\midrule
\multirow{3}{*}{ResNet50}   & Standard         & \best 20.24             \\
                            & CLIP             & 12.71             \\
                         
                            & CLIP-Auto        &  12.92             \\ \midrule
\multirow{3}{*}{ResNet101}  & Standard         & \best24.25             \\
                            & CLIP             & 17.01             \\
                           
                            & CLIP-Auto        & 17.20             \\ \midrule
\multirow{3}{*}{ViT-B/32}   & Standard         & \best 30.58             \\
                            & CLIP             & 17.19             \\
                           
                            & CLIP-Auto        & 16.80             \\ \midrule
\multirow{3}{*}{ViT-B/16}   & Standard         & \best 36.36            \\
                            & CLIP             &  25.76             \\
                           
                            & CLIP-Auto        &  25.86         \\ \midrule

\multirow{2}{*}{ResNet50x4} & CLIP             & 21.32             \\
                           
                            & CLIP-Auto        & \best21.43             \\ \midrule

\multirow{2}{*}{ResNet50x16} & CLIP             &\best 20.52             \\
                         
                            & CLIP-Auto        &  20.07            \\ \bottomrule

\end{tabular}
}
    \caption{{ Model accuracies on Stylized ImageNet. CLIP is not able to improve the robustness performance over the corresponding standard models. CLIP and CLIP-Auto perform similarly.}}
    \label{tab:stylized}
\end{table}

\subsection{Distribution Shifts}
\paragraph{Natural Distribution Shifts} Performance per dataset in natural distribution shifts is shown in Figure~\ref{fig:ndistbreakdown}. The CLIP results are compared to the results of standard ImageNet models in the testbed~\cite{taori2020measuring}. We find that CLIP models obtain state-of-the-art effective robustness on all datasets. Compared to CLIP models, CLIP-Auto obtains comparable performance. The reason is that the pre-trained image features are the key to the performance, while the prompts that only synthesize the classifiers based on the image features have limited impact.

\begin{table*}[]
\resizebox{\linewidth}{!}{
\begin{tabular}{llccccccccccccccccc}
\hline 
                                 \multirow{2}{*}{\textbf{Model}}  &                  & \multicolumn{1}{l}{}  & \multicolumn{1}{l|}{}                 & \multicolumn{3}{c|}{\textbf{Noise}}                                    & \multicolumn{4}{c|}{\textbf{Blur}}                                                       & \multicolumn{4}{c|}{\textbf{Weather}}                                                & \multicolumn{4}{c}{\textbf{Digital}}                                  \\
                                 &                  & \textbf{Original Acc} & \multicolumn{1}{c|}{\textbf{Avg Acc}} & \textbf{Gauss} & \textbf{Shot} & \multicolumn{1}{c|}{\textbf{Impulse}} & \textbf{Defocus} & \textbf{Glass} & \textbf{Motion} & \multicolumn{1}{c|}{\textbf{Zoom}} & \textbf{Snow} & \textbf{Frost} & \textbf{Fog} & \multicolumn{1}{c|}{\textbf{Bright}} & \textbf{Contrast} & \textbf{Elastic} & \textbf{Pixel} & \textbf{JPEG} \\ \hline
\multirow{3}{*}{ResNet50}   & Standard         & \best 76.13                 & \best 39.17                & \best 29.29          &\best 27.03         & \best 23.81            & \best 38.75            & \best 26.79          & \best 38.67           & \best 36.24         & \best 32.53         & \best 38.14          & \best 45.83        & \best 68.02           & \best39.06             & \best45.25            &\best 44.79          & \best53.41         \\
                                 & CLIP             & 59.85                 & 26.67                & 19.63          & 18.09         &  12.23            & 26.06            & 14.78          & 25.36           & 23.61         & 22.68         & 26.98          & 38.55        & 50.35           & 32.89             & 28.29            & 28.13          & 32.42         \\
                                 
                                 & CLIP-Auto         &  61.78                 &  27.46                &  19.93          &  18.31         & 11.83            &  27.09            &  14.90          &  26.29           &  24.02         &  23.26         &  27.82          &  39.46        &  52.38           & 33.95             & 29.08            & 29.58          & 33.97         \\ \hline
\multirow{3}{*}{ResNet101}  & Standard         & \best 77.37                 & \best 44.10                & \best 34.77          & \best 32.63         & \best 29.02            & \best 44.27            & \best32.60           &\best 44.31           & \best40.78         &\best 36.49         &\best 41.73          &\best 49.14        & \best69.85           &\best 42.98             & \best50.11            &\best 53.51          & \best59.30         \\
                                 & CLIP             & 62.32                 & 32.07                & 24.81          & 23.34         &  19.03            & 31.17            &  18.77          & 30.14           & 26.90         &  28.32         & 31.53          & 43.38        & 54.57           & 39.01             & 31.69            & 38.91          & 39.49         \\
                                
                                 & CLIP-Auto         & 63.83                 & 32.57                & 25.31          & 23.60         & 18.93            & 31.81            & 18.75          & 30.85           & 27.51         & 28.20         & 31.80          & 44.14        & 55.68           & 39.89             & 32.11            & 39.44          &  40.46         \\ \hline

\multirow{3}{*}{ViT-B/32}   & Standard         & \best 78.73                 & \best 50.79                & \best 44.05          &\best  41.40         &\best 41.30            &\best 48.59            & \best45.81          &\best 54.48           & \best43.20         & \best31.50         &\best 38.99          & \best56.81        & \best67.77           & \best59.72             & \best57.45            & \best66.59          & \best64.13         \\
                                 & CLIP             & 63.40                 & 40.31                & 37.21          & 35.36         & 33.65            & 40.34            & 29.18          & 41.17           & 32.13         & 34.29         & 37.13          & 46.52        & 57.62           & 45.22             & 40.96            & 47.26          & 46.60         \\
                                 
                                 & CLIP-Auto         & 65.16                 & 41.04                & 37.77          & 35.69         & 33.95            & 41.12            & 29.44          & 42.19           & 32.65         & 34.65         & 37.43          & 47.27        & 59.14           & 46.15             & 41.97            & 48.35          & 47.81         \\ \hline
\multirow{3}{*}{ViT-B/16}   & Standard         & \best 84.20                & \best60.12 & \best54.11 & \best52.74 & \best52.14 & \best54.28 & \best51.4 & \best62.84 &\best 55.21 & \best60.48 & \best59.03 & \best61.48 & \best78.18 & \best55.70 & \best63.15 & \best71.46 &\best 69.63         \\
                                 & CLIP             & 68.43                 & 43.87 & 39.48 & 37.82 & 35.05 & 42.03 &  33.22 & 44.75 & 36.60 & 43.52 & 42.72 & 51.53 & 62.33 & 47.92 & 41.27 & 49.76 & 50.01         \\
                                
                                 & CLIP-Auto         & 69.51                 &  44.19 & 39.79 & 38.09 & 35.08 & 42.44 & 33.17 & 45.08 & 36.83 & 43.93 & 42.94 & 51.76 & 63.17 & 48.09 & 41.30 & 50.24 & 50.91       \\ \hline

\multirow{2}{*}{ResNet50x4} & CLIP             & 66.28                 & \best34.43                & \best27.76          & \best26.57         & \best23.59            &\best31.36            & \best18.70           &\best 31.67           & \best28.34         & \best30.77         &\best 34.39          &\best 45.72        & 57.81           &\best 39.80             & \best32.57            & 42.40           & 44.94         \\
                                
                                 & CLIP-Auto         & \best66.48                 & 33.98                & 27.05          & 25.82         & 22.84            & 30.60            & 18.04          &  30.85           & 27.75         & 30.44         & 33.98          & 45.15        &\best 57.83           & 39.78             & 32.21            &  42.17          & 45.23         \\ \hline
                                 
\multirow{2}{*}{ResNet50x16} & CLIP             & 70.67                 & \best 41.39 & \best 38.22 & \best 37.16 & \best 35.15 & \best 36.60 & \best 23.49 & \best 38.34 & \best 34.08 & \best 38.08 & \best 39.33 & \best 50.88 & \best 62.37 & \best 46.23 & \best 37.68 & 49.67 & 53.63         \\

                                 & CLIP-Auto         & \best 70.81                & 40.93 & 37.63 & 36.34 & 34.68 & 36.22 & 22.55 & 37.55 & 32.95 & 37.22 & 38.93 & 50.56 & 62.42 & 46.00 & 37.13 & \best 50.05 & \best 53.66  \\ \hline

\end{tabular}

}
\caption{{ Individual top-1 accuracy scores on all the corruption types of ImageNet-C. ``Original Acc'' refers to the accuracy of the clean ImageNet validation set. ``Avg Acc'' denotes the average accuracy of 15 common corruptions.} }
\label{tab:inc}
\end{table*}


\begin{table*}[]
\begin{subtable}{\linewidth}\centering
\resizebox{1\linewidth}{!}
{
\begin{tabular}{llccccccccccc}
\hline
\multicolumn{2}{l}{\textbf{}}                       & \multicolumn{1}{r|}{\textbf{}}    & \multicolumn{2}{c|}{\textbf{Noise}}                    & \multicolumn{2}{c|}{\textbf{Blur}}                   & \multicolumn{2}{c|}{\textbf{Weather}}                & \multicolumn{4}{c}{\textbf{Digital}}                                  \\
\multicolumn{2}{l}{\textbf{Model}}                  & \multicolumn{1}{c|}{\textbf{mFR}} & \textbf{Gauss} & \multicolumn{1}{c|}{\textbf{Shot}} & \textbf{Motion} & \multicolumn{1}{c|}{\textbf{Zoom}} & \textbf{Snow} & \multicolumn{1}{c|}{\textbf{Bright}} & \textbf{Translate} & \textbf{Rotate} & \textbf{Tilt} & \textbf{Scale} \\ \hline

\multirow{3}{*}{ResNet50}   & Standard         & \best 57.90                         &\best 59.00                & \best58.00                & \best64.00                &\best 72.00                & \best63.00                & \best62.00                & \best44.00                & \best52.00                & \best57.00                & \best48.00                \\
                                 & CLIP             & 113.69                        & 99.59                & 93.34                & 129.19               & 147.51               & 118.32               & 148.67               & 99.13                & 102.11               & 129.09               & 70.00                \\
                                
                                 & CLIP-Auto        &  110.62                        &  95.18                &  90.29                &  125.63               &  143.40               &  117.65               &  143.62               &  96.41                &  99.66                &  125.83               &   68.55                \\ \hline
\multirow{3}{*}{ResNet101}  & Standard         & \best  53.02                         & \best55.10                & \best51.70                &\best 53.77                & \best63.25                & \best58.80                &\best 57.32                & \best42.25                & \best48.96                &\best 53.65                & \best45.35                \\
                                 & CLIP             & 98.91                         & 82.82                & 78.80                & 108.99               & 129.43               & 104.07               & 129.08               & 89.22                & 90.17                & 113.72               & 62.79                \\
                                
                                 & CLIP-Auto        &  97.23                         &  81.51                &  77.38                &  108.10               &  127.41               &  103.15               &  127.22               &  86.67                &  88.12                &  110.71               &  62.08                \\ \hline
\multirow{3}{*}{ViT-B/32}   & Standard         & \best 35.26                         & \best 28.23                & \best27.76                &\best 32.70                & \best46.46                & \best34.86                &\best 56.44                & \best35.79                & \best45.69                & \best43.05                & \best44.54                \\
                                 & CLIP             & 74.53                         & 64.56                & 58.40                & 66.85                & 93.94                & 71.13                & 94.87                & 67.26                & 76.61                & 88.34                & 63.35                \\
                               
                                 & CLIP-Auto        &  72.14                         &  61.17                & 56.20                & 65.02                & 92.26                & 70.00                & 90.61                & 64.49                & 74.81                & 85.04                &  61.81                \\ \hline
\multirow{3}{*}{ViT-B/16}   & Standard         &\best 33.15&\best34.13&\best33.86&\best27.17&\best39.63&\best19.90&\best47.63&\best26.03&\best31.84&\best36.30&\best35.06        \\
                                 & CLIP             &64.85&57.37&52.92&60.26&85.93&54.98&79.75&57.62&67.08&79.96&52.59           \\
                           
                                 & CLIP-Auto        &  63.58&56.11&52.03&59.67&83.72&54.28&77.24&56.95&66.13&78.00&51.72      \\ \hline

\multirow{2}{*}{ResNet50x4} & CLIP             &  90.07                         &\best  71.05                &\best  68.37                & 107.20               &  125.19               & \best101.92               &  117.69               &  72.61                & 79.64                &  101.59               &  55.47                \\
                               
                                 & CLIP-Auto        & \best  89.42                         &  71.15                &  68.57                & \best 106.49               & \best 122.51               &   102.64               & \best 116.73               & \best 72.52                & \best 78.80                & \best  99.77                & \best  55.03                \\ \hline

\multirow{2}{*}{ResNet50x16} & CLIP             & 76.78&57.59&55.10&92.47&111.45&\best88.87&102.99&56.70&67.19&89.35&46.08                \\
                              
                                 & CLIP-Auto        &\best76.23&\best56.67&\best54.61&\best92.22&\best110.02&89.08&\best101.74&\best56.51&\best67.06&\best88.75&\best45.60     \\ \hline
\end{tabular}
}

\caption{{ The mean flip rate (mFR) across all perturbations.}}
\label{tab:imgnetPmfr}
\end{subtable}

\begin{subtable}{\linewidth}\centering
\resizebox{1\linewidth}{!}
{
\begin{tabular}{llccccccccccc}
\hline
\multicolumn{2}{c}{\textbf{}}                       & \multicolumn{1}{c|}{\textbf{}}     & \multicolumn{2}{c|}{\textbf{Noise}}                    & \multicolumn{2}{c|}{\textbf{Blur}}                   & \multicolumn{2}{c|}{\textbf{Weather}}                 & \multicolumn{4}{c}{\textbf{Digital}}                                 \\
\multicolumn{2}{l}{\textbf{Model}}                  & \multicolumn{1}{c|}{\textbf{mT5D}} & \textbf{Gauss} & \multicolumn{1}{c|}{\textbf{Shot}} & \textbf{Motion} & \multicolumn{1}{c|}{\textbf{Zoom}} & \textbf{Snow} & \multicolumn{1}{c|}{\textbf{Bright}} & \textbf{Translate} & \textbf{Rotate} & \textbf{Tilt} & \textbf{Scale} \\ \hline

\multirow{3}{*}{ResNet50}   & Standard         & \best 78.20         &  \best82.00              &  \best79.00         &  \best84.00            &  \best89.00          & \best 80.00            & \best 84.00             &  \best64.00              &  \best 73.00           &  \best80.00         &  \best67.00          \\
                                 & CLIP             & 113.36        & 100.52             & 94.79         & 128.12           & 144.02         &   118.11           & 144.64            & 97.28              & 102.42          & 124.14        & 79.60          \\
                                 
                                 & CLIP-Auto  &   112.58        &  99.03              &  94.11         &  126.98           &  142.5          &  118.68           &  142.34            &  97.22              &  102.13          &  123.32        &   79.52          \\ \hline
\multirow{3}{*}{ResNet101}  & Standard         & \best 74.37         &\best  80.78              & \best 75.62         & \best 74.82            & \best 81.54          & \best 77.05            &\best  79.30             &\best  62.45              & \best  70.72           & \best 76.21         & \best 65.22          \\
                                 & CLIP             & 102.95        & 88.59              & 84.83         & 113.54           & 130.41         &  107.79           & 130.73            & 91.34              & 94.19           & 113.95        & 74.17          \\
                                 
                                 & CLIP-Auto  &  102.60        & 88.13              & 84.42         & 113.51           & 129.98         & 107.99           & 130.05            & 90.82              & 93.75           & 113.49        &  73.87          \\ \hline
\multirow{3}{*}{ViT-B/32}   & Standard         & \best 66.58         &  \best42.42              &  \best42.20         &  \best45.85            & \best 56.53          & \best 45.55            &  \best69.09             &  \best 189.21             &  \best 59.54           &  \best 55.92         &  \best 59.53          \\
                                 & CLIP             & 81.19         & 72.78              & 66.85         & 76.38            & 98.42          & 78.51            & 99.52             & 71.46              & 82.40           & 91.56         & 74.04          \\
                                 
                                 & CLIP-Auto  &  80.04         &  70.46              &  66.04         &  75.39            &  97.20          &  77.84            &  96.97             &  70.83              &  81.85           &  90.29         &  73.54          \\ \hline
\multirow{3}{*}{ViT-B/16}   & Standard         & \best  53.26&\best58.81&\best57.83&\best45.97&\best56.56&\best33.92&\best68.96&\best43.05&\best52.33&\best60.05&\best55.17      \\
                                 & CLIP             &  77.13& 70.69& 65.90& 73.60& 94.08& 67.82&91.99& 68.16& 79.38& 91.22& 68.49      \\
                                
                                 & CLIP-Auto  &77.17& 70.20&66.06&73.85&93.90&68.25&90.79&68.67&79.74&91.53&68.72      \\ \hline

\multirow{2}{*}{ResNet50x4} & CLIP             &  98.25         &  80.76              &  78.25         &  114.14           & 128.89         & \best 109.35           &  124.89            &  80.48              &  88.09           &  108.25        &  69.42          \\
                                
                                 & CLIP-Auto &  \best 97.74         & \best 80.19              & \best 77.62         &\best  113.99           & \best 127.88         &  110.22           &\best  123.88            & \best 79.87              &\best  87.47           & \best 107.30        & \best 69.00          \\ \hline
\multirow{2}{*}{ResNet50x16} & CLIP             &  90.52&71.21&68.30&107.04&123.43&\best102.17&117.19&69.26&80.68&103.20&62.69       \\
                                 
                                 & CLIP-Auto &  \best 89.73&\best 70.28&\best 67.59&\best 106.36&\best 122.45& 102.30&\best 115.45&\best 68.37&\best 80.15&\best 102.08&\best  62.30      \\ \hline

\end{tabular}
}
\caption{{ The mean top-5 distance (mT5D) across all perturbations.}}
\label{tab:imgnetPmt5d}
\end{subtable}
\caption{{ Results on ImageNet-P. CLIP reduces the robustness compared to standard models.}}
\label{tab:imagenetpbreakdown}
\end{table*}

\paragraph{Synthetic Distribution Shifts} We show the performance on each individual dataset of synthetic distribution shifts. Figure~\ref{fig:imagenetCbreakdown} illustrates that CLIP models fail to improve the robustness over corresponding standard ImageNet models on ImageNet-C, which is in contrast to the results on natural distribution shifts. This also confirms the findings in \cite{taori2020measuring} that robustness under synthetic distribution shifts does not imply that the corresponding model has robustness on natural distribution shifts. Similar to the observation on natural distribution shifts, CLIP-Auto achieves comparable effective and relative robustness with CLIP, suggesting that the impact of CLIP models on the robustness is limited.

In Table~\ref{tab:imagenetpbreakdown}, we compare two common metrics on ImageNet-P: mean flip rate (mFR) and mean top-5 distance (mT5D). We find that all CLIP and CLIP-Auto models generally reduce the performance compared to the standard models. 
In addition, we show the results on Stylized ImageNet in Figure~\ref{fig:stylizedimagenetbreakdown} and Table~\ref{tab:stylized}. We draw the same conclusion that CLIP does not improve robustness. CLIP and CLIP-Auto produce comparable performance.

\begin{figure*}
\centering
\subcaptionbox{{ ImageNet.}\label{fig:adv_robustness_imgNet}}{\includegraphics[width=0.45\textwidth]{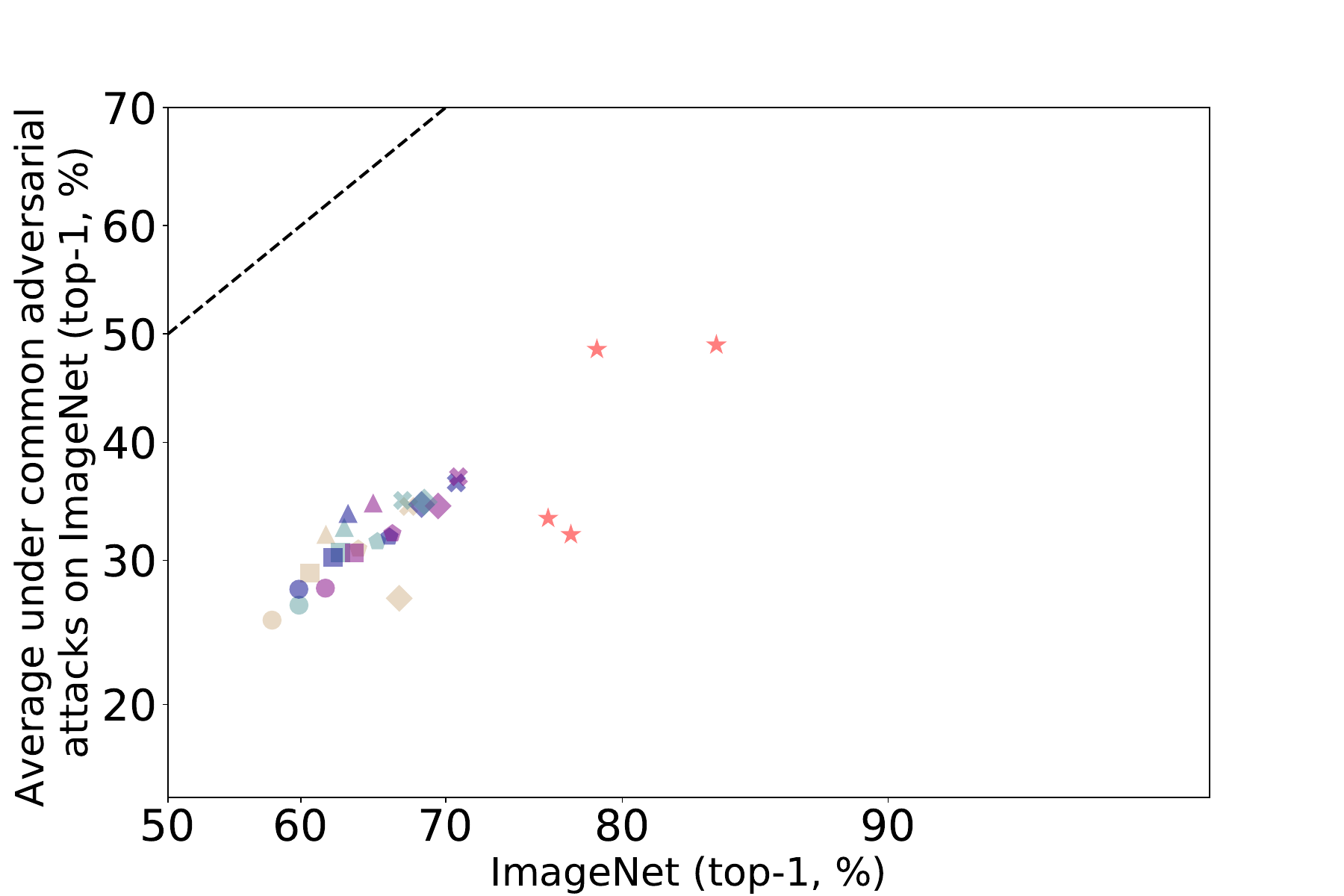}}
\hspace{0.02in}
\subcaptionbox{{ CIFAR-10.}\label{fig:adv_robustness_cifar}}{\includegraphics[width=0.45\textwidth]{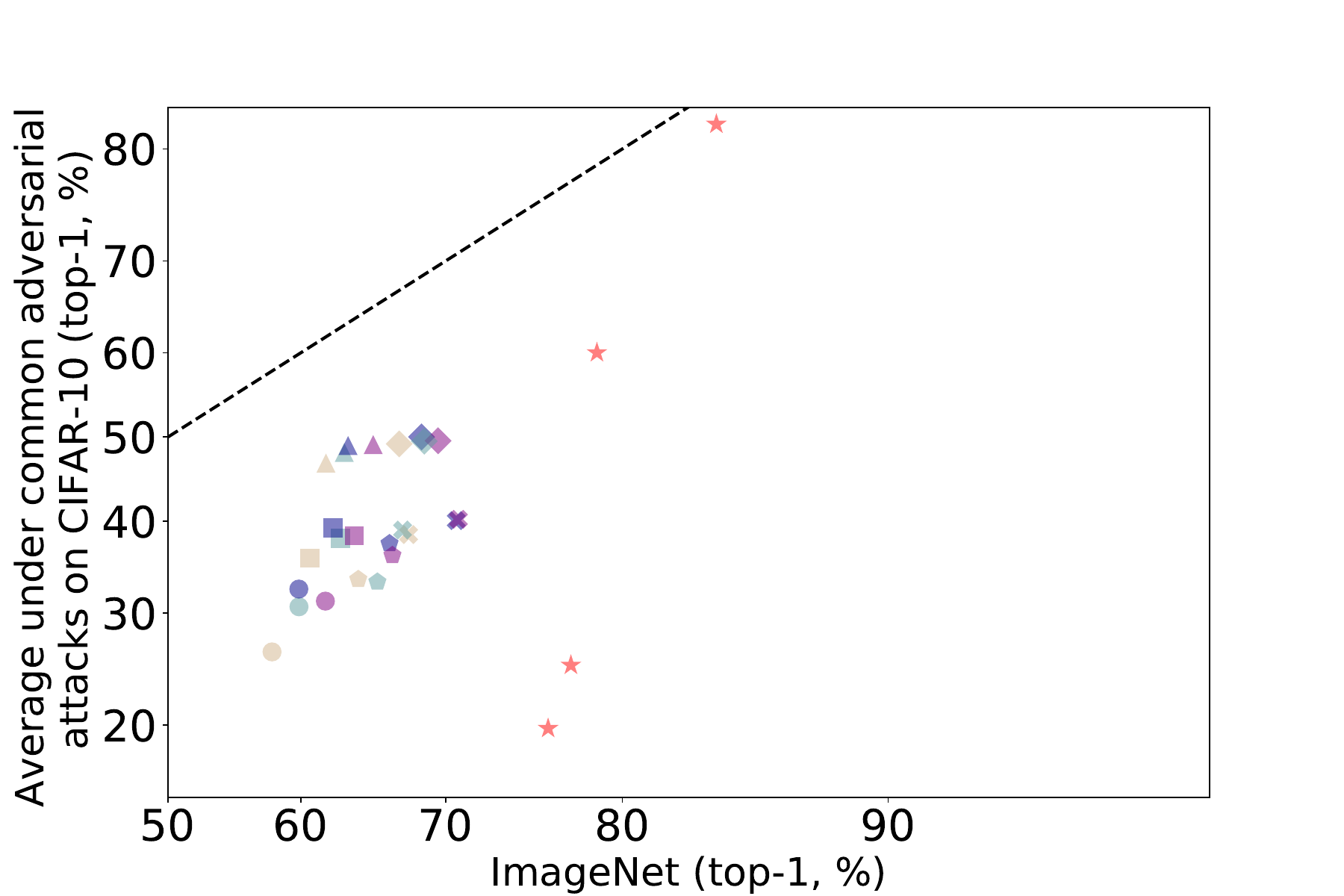}}
\\
\includegraphics[width=0.95\linewidth]{figs/legend.pdf}%
\caption{{ Model accuracies under common adversarial attacks on ImageNet and CIFAR-10. Similar to the results on synthetic distribution shifts, CLIP is more vulnerable to adversarial attacks than standard models. Red:  standard ImageNet models.  Blue:  zero-shot CLIP models.  Purple: CLIP-Auto models.}}
\label{fig:advbreakdown}
\end{figure*}

\begin{table*}[]
\resizebox{\linewidth}{!}{
\begin{tabular}{llcccccccccc}
\hline
\multicolumn{2}{l|}{\textbf{Attack Setting}  }                               & \multicolumn{4}{c|}{\textbf{White-Box Attacks}   }                                                                                                                               & \multicolumn{4}{c|}{\textbf{Transfer-Based Attacks}}                                                                                                                                  & \multicolumn{2}{c}{\textbf{Black-Box Attacks}    }                                        \\
\multicolumn{2}{l|}{\textbf{Model}}                                          & \textbf{FGSM}        & \textbf{DeepFool}    & \textbf{BIM}     & \multicolumn{1}{c|}{\textbf{MIM}} & \textbf{FGSM}         & \textbf{BIM}         & \textbf{MIM}          & \multicolumn{1}{c|}{\textbf{DIM}} & \textbf{NES}          & \textbf{SPSA}          \\ \hline
\multirow{3}{*}{ResNet50}   & Standard                              & 0.001 / 34.60                          & 0.0001  / 0                           & 0.001 / 0                            & 0.001 / 0                        & 0.019 / 44.74                         & 0.026 / 42.16                          & 0.025 / 36.00                          & 0.027 / 37.74                         & 0.001 / 1.10                           & 0.001 / 1.10                           \\
                            & CLIP                                  &  0.001 / 35.30  & 0.0002 / 0                            & 0.001 /  0                           & 0.001 /  0                       & \best 0.086 / 61.40    & \best 0.138 / 61.80     & \best 0.103 /  60.00    & \best 0.115 /  70.20   &  0.002 / 17.70  & \best 0.002 / 18.40     \\
                          
                            & CLIP-Auto         & \best 0.001 / 36.40     & 0.0002 / 0                            & 0.001 / 0                            & 0.001 / 0                        &  0.081 / 61.30 &  0.136 / 61.30  &   0.101 / 59.30 &  0.116 / 59.20 & 0.002 / 17.10                          &   0.002 / 17.60 \\ \hline
\multirow{3}{*}{ResNet101}  & Standard                              & 0.001 / 43.13                          & 0.0002 / 0                            & 0.001 / 0                            & 0.001 / 0                        & 0.025 / 56.36                         & 0.041 / 50.27                          & 0.040 / 45.90                          & 0.047 / 49.03                         & 0.002 / 3.00                           & 0.002 / 2.40                           \\
                            & CLIP                                  & \best 0.001 / 45.70     & 0.0004 / 0                            & 0.001 / 0                            & 0.001 / 0                        & \best 0.225 / 69.60    & \best 0.281 /  70.30    & \best 0.174 /  69.40    & \best 0.220 /  70.20   & \best 0.004 / 33.90     & \best 0.005 / 33.10     \\
                         
                            & CLIP-Auto         &  0.001 / 44.30  & 0.0004 / 0                            & 0.001 / 0                            & 0.001 / 0                        &  0.222 / 68.50 &  0.274 / 69.20  &  0.189 / 68.20  &  0.227 / 68.40 & 0.005 /  32.40                         & 0.005 / 32.50                          \\ \hline
\multirow{3}{*}{ViT-B/32}   & Standard                              & \best 0.016 / 37.60     & \best 0.0106 / 9.40    & \best 0.006 / 0.80    & \best 0.007 / 0.6 & \best 0.576 / 94.10    & \best  0.876 / 94.60    & \best0.591 / 94.10      & \best0.594 / 94.30     & \best 0.093 / 87.30     & \best 0.094 / 87.30     \\
                            & CLIP                                  & 0.005 / 26.79                          &  0.0020 / 0.25 & 0.002 / 0                            & 0.002 / 0                        & 0.692 / 83.98                         &   0.870 / 88.30 & 0.582 / 83.90                          & 0.592 / 84.12                         & 0.023 / 61.10                          &  0.023 / 61.30  \\
                          
                            & CLIP-Auto         & 0.007 / 26.24                          & 0.0023 /  0.10                        & 0.002 / 0.06                         & 0.002 / 0                        &  0.687 / 84.23 & 0.875 / 87.23                          &  0.583 / 84.18  &  0.592 / 86.00 &   0.024 / 61.60 & 0.023 / 61.20                          \\ \hline

\multirow{3}{*}{ViT-B/16}   & Standard                              & \best - / 91.30&
0.0026 / 0&
\best- / 91.10&
\best- / 91.20&
\best- / 93.80&
\best- / 95.10&
\best- / 94.30&
\best- / 93.90&
\best- / 83.90&
\best- / 83.80   \\
                            & CLIP                                  & 0.004 / 30.20& 
0.0020 / 0& 
0.002 / 0& 
0.002 / 0& 
0.5005 / 84.80& 
0.573 / 88.20& 
0.454 / 85.20& 
0.503 / 85.40& 
0.024 / 62.80& 
0.025 / 63.50  \\
            
                            & CLIP-Auto         & 0.005 / 30.10&
0.0030 / 0&
0.002 / 0 &
0.002 / 0&
0.543 / 84.40&
0.573 / 87.80&
0.464 / 85.30&
0.5065 / 85.30&
0.026 / 61.30&
0.026 / 61.30                        \\ \hline
    
\multirow{2}{*}{ResNet50x4} & CLIP                                  & \best 0.001 / 54.00     & 0.0003 / 0                            & 0.001 / 0                            & 0.001 / 0                        & \best 0.274 / 63.20    & \best 0.312 / 64.80     & \best 0.262 /  62.70    & \best 0.301 /  61.50   & \best 0.004 / 34.90     &   0.004 / 33.50 \\
                           
                            & CLIP-Auto         &  0.001  / 50.90 & 0.0003 / 0                            & 0.001 / 0                            & 0.001 / 0                        &  0.266 / 60.70 &  0.306 / 63.40  &  0.256 / 59.60  &  0.275 / 60.30 &  0.004 / 32.90  & \best 0.005 / 33.60     \\ \hline
               
\multirow{2}{*}{ResNet50x16} & CLIP                                  &  0.001 / 62.80& 
0.0003 / 0& 
0.001 / 0& 
0.001 / 0& 
0.443 / 69.60& 
0.451 / 68.10& 
0.342 / 68.80& 
0.408 / 68.40& 
\best0.004 / 31.40& 
\best0.004 / 31.70 \\
                           
                            & CLIP-Auto         & \best0.001 / 62.90&
0.0003 / 0&
0.001 / 0&
0.001 / 0&
\best0.447 / 69.80&
\best0.494 / 69.00&
\best0.361 / 69.20&
\best0.404 / 69.50&
0.006 / 30.40&
0.006 / 31.40     \\ \hline
                            
\end{tabular}

}

\caption{{ Model results against individual untargeted adversarial attacks under the $l_{\infty}$ norm on CIFAR-10. Each entry consists of the median $l_{\infty}$ distance of the minimum adversarial perturbations over all samples on the left, and the model accuracy for the perturbation budget $\epsilon = 8/255$ on the right. }}
\label{tab:adv}
\end{table*}

\subsection{Adversarial Attacks}
We briefly provide some additional numerical details of the zero-shot CLIP models under adversarial attacks.

\paragraph{Common Attacks} In Figure~\ref{fig:advbreakdown}, we see that all CLIP models obtain comparable average effective robustness with the standard models on both ImageNet and CIFAR-10. However, we see little evidence of relative robustness improvements. In particular, we find that the robustness drop on ImageNet is more significant than on CIFAR-10. The reason is that the gains of the standard models are mainly from the similar distribution in ImageNet. Note that on CIFAR-10, we use CLIP in a zero-shot manner, and use CLIP-Auto and linear-probe standard models for comparison. In Table~\ref{tab:adv}, we show detailed results under each adversarial attack on CIFAR-10. We also show the median distance of the minimum adversarial perturbations. We find that the CLIP is more vulnerable under transfer-based attacks and black-box attacks than standard models. CLIP-Auto does not make much difference in the performance compared to CLIP.

\paragraph{Typographic Attacks} CLIP consists of multimodal neurons which respond to both images and text for a given concept. Therefore this particular type of attack~\cite{goh2021multimodal} is designed for zero-shot CLIP models. As all the CLIP models are built based on classifiers from the text, CLIP can be very vulnerable to such attacks. In Figure~\ref{fig:typobreakdown}, we compare the model accuracies on our new robustness datasets: ImageNet-T and CIFAR-10-T. Unsurprisingly, all CLIP models are vulnerable to typographic attacks and cause significant robustness drop compared to standard models. We find that typographic attacks reduce both the ImageNet in-distribution and out-of-distribution (ImageNet-T and CIFAR-10-T) performance by a large amount compared to the standard models. We show that the success rate is also much higher for the CLIP models. CLIP-Auto models are not able to significantly improve the robustness of the CLIP models. More details of ImageNet-T and CIFAR-10-T are described in Appendix~\ref{sec:imagentcifart}.

\section{The ImageNet-T and CIFAR-10-T Robustness Test Sets}
\label{sec:imagentcifart}
We use typographic attacks to create ImageNet-T and CIFAR-10-T.

\begin{figure*}
\centering
\subcaptionbox{{ ImageNet-T.}}{\includegraphics[width=0.95\textwidth,height=0.60\textwidth]{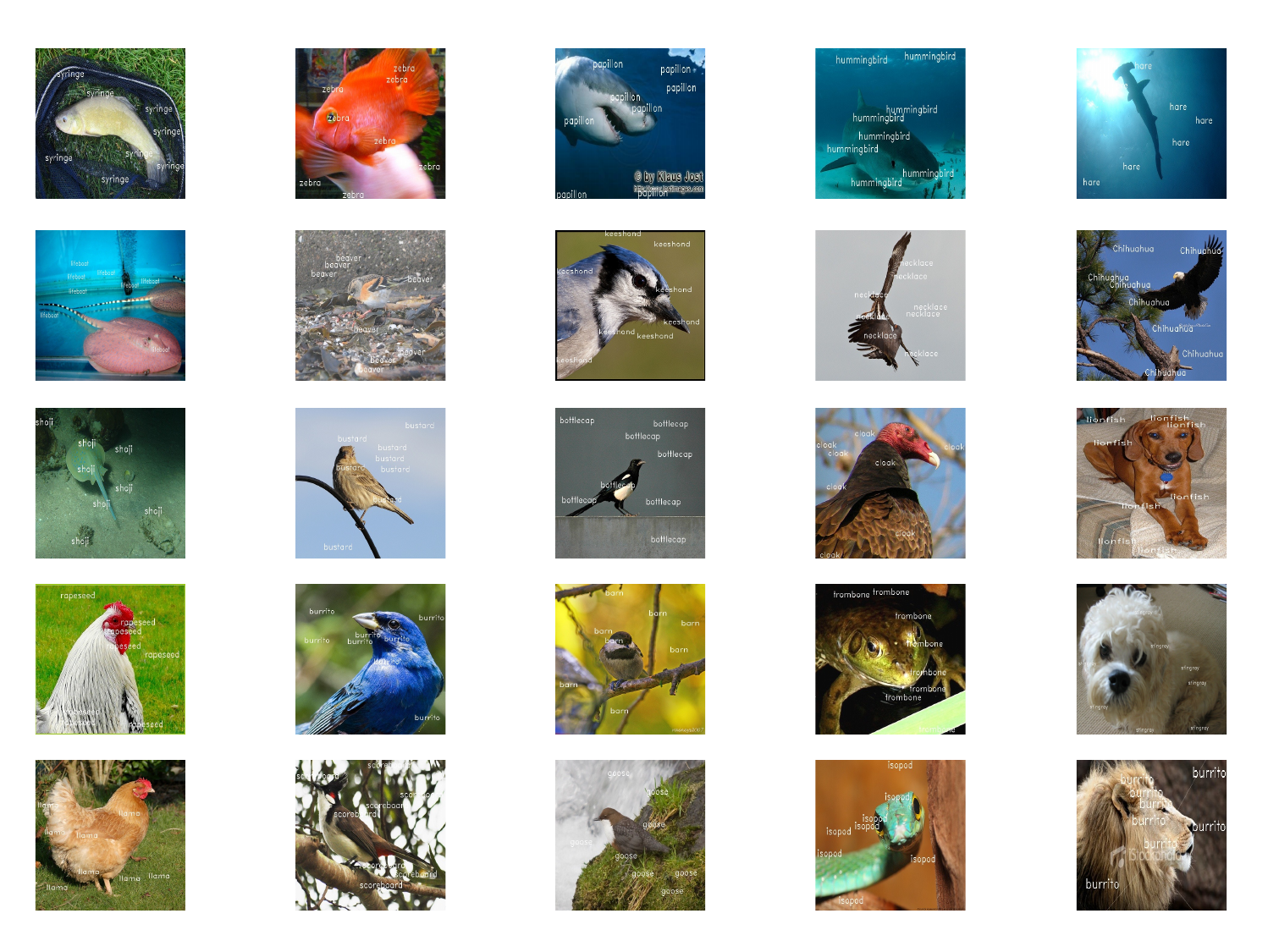}}
\hspace{0.02in}
\subcaptionbox{{ CIFAR-10-T.}}{\includegraphics[width=0.95\textwidth,height=0.60\textwidth]{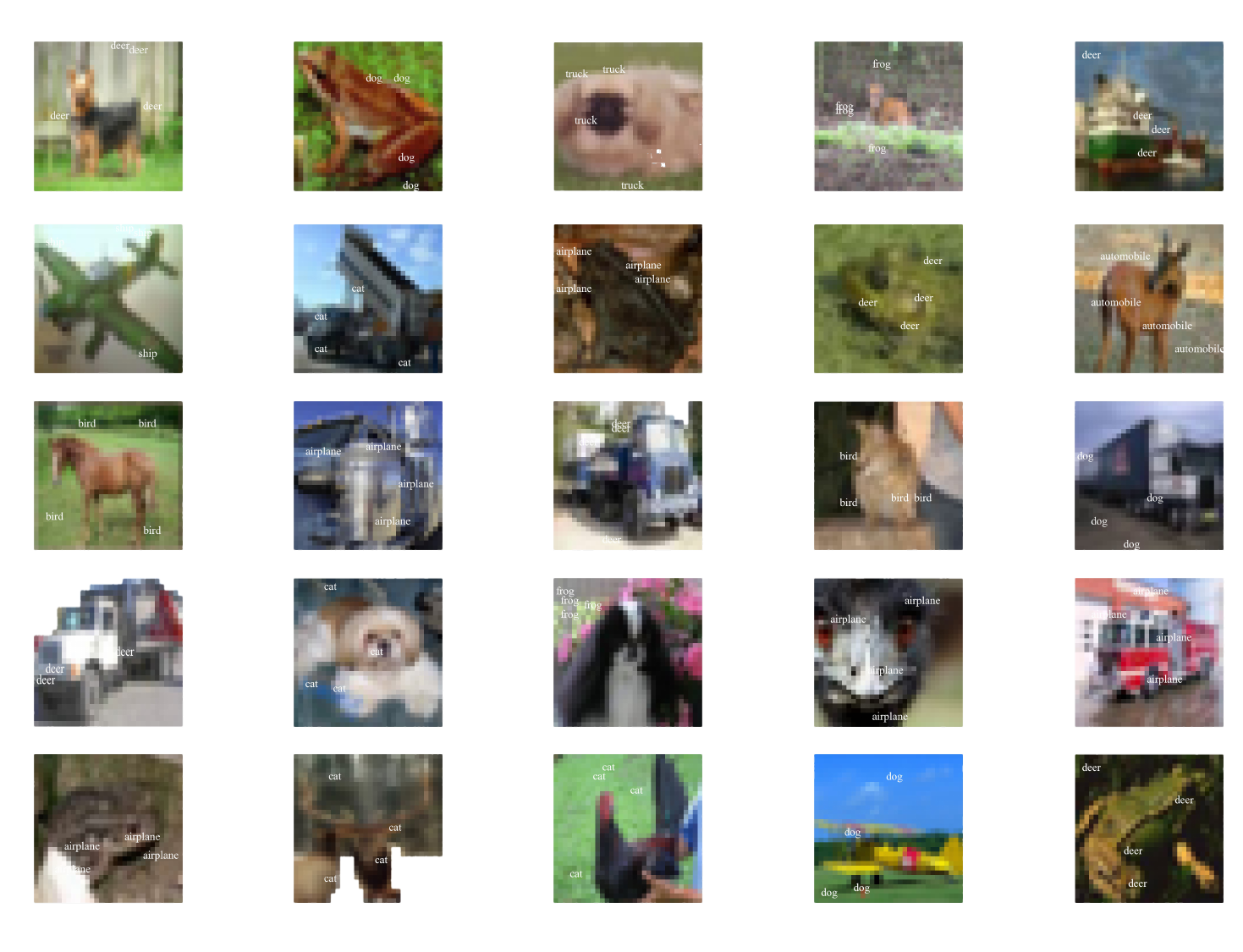}}
    \caption{{ Our ImageNet-T and CIFAR-10-T samples. The attack text for each image of ImageNet-T and CIFAR-10-T is a uniformly sampled label text from other classes in ImageNet and CIFAR-10 respectively.}}
    \label{fig:imagenet-cifar-t}
\end{figure*}

\begin{figure*}[h]
    \centering
    \includegraphics[width=0.75\textwidth]{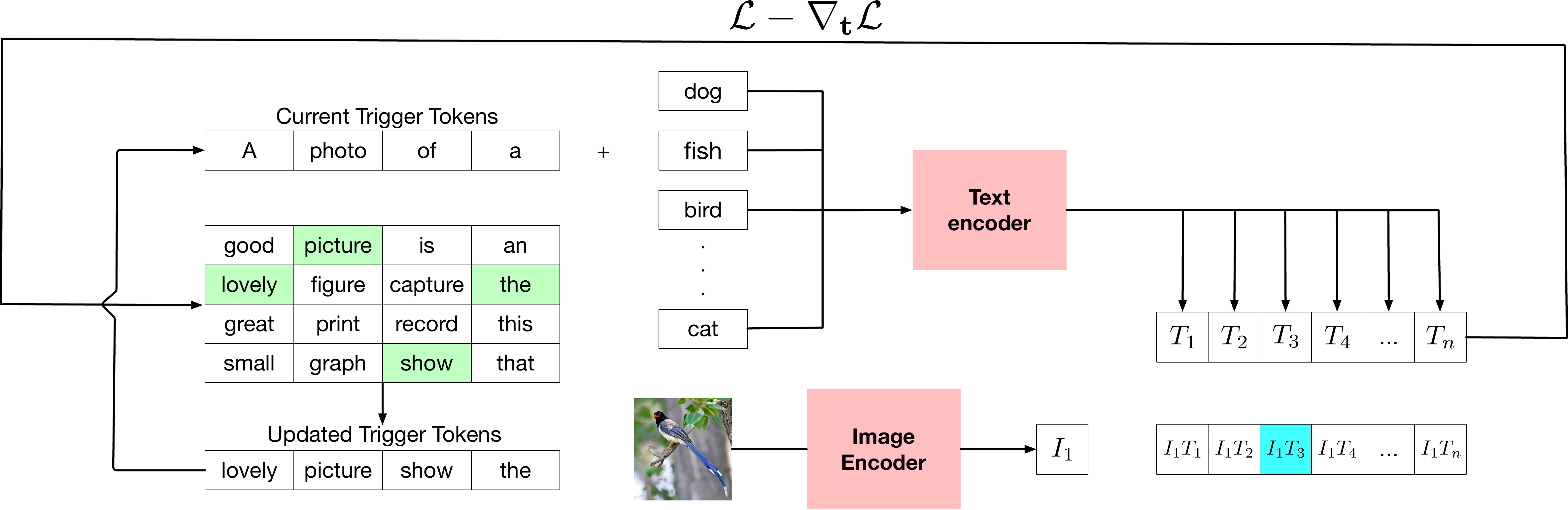}
    \vspace{-0.05in}
    \caption{{Summary of producing an automated prompt for CLIP-Auto. A prompt is created using a template that combines a set of trigger tokens with the label text. The trigger tokens are shared across all classes and decided via a gradient-based search. At each search step, we use the current trigger tokens and each label text to synthesize a linear classifier, then compute the gradients for candidate trigger tokens and update the trigger tokens with that lead to the smallest loss. After iteratively repeating this process, the trigger tokens converge and are returned to create a prompt.}}
    \label{fig:overview}
\end{figure*}
\paragraph{ImageNet-T Design} We aim to quantitatively evaluate the image classification robustness under the typographic attacks~\cite{goh2021multimodal}. Attacks are automatically generated using the same (arbitrarily chosen) eight coordinates and using a consistent font style, as shown in Figure~\ref{fig:imagenet-cifar-t}. As the setup is the targeted attack, we consider an attack to have succeeded if the predicted class is changed to the attack class. The attack text for each image is a target label text uniformly chosen over other classes except its true class at random. We use OpenCV\footnote{{\small\url{https://opencv.org/}}} to attach the attack text to the images. The dataset contains 50,000 images, which equals the size of the ImageNet validation set. As documented in \cite{goh2021multimodal}, the idea of typographic attacks is in general similar to work such as adversarial patches~\cite{brown2017adversarial} and physical adversarial examples~\cite{athalye2018synthesizing}. We plan to investigate more attacks along this line as one of the future directions.

\paragraph{CIFAR-10-T Design} We aim to provide a small typographic attacks based test set for quicker experimentation. The only difference from the ImageNet-T is that we use four coordinates instead of eight due to the lower resolution of the images in CIFAR-10. This results in 10,000 images in CIFAR-10-T, which is the size of the CIFAR-10 test set.

We show image examples of ImageNet-T and CIFAR-10-T in Figure~\ref{fig:imagenet-cifar-t}.

\section{Data Overlap Examples}
\begin{figure*}[]
\centering
\subcaptionbox{{\small ImageNetV2. }\label{fig:imagenet-clean}}{\includegraphics[width=0.85\textwidth]{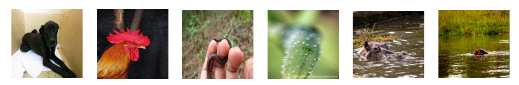}}
\\
\subcaptionbox{{\small Google Images. }\label{fig:imagenet-typo}}{\includegraphics[width=0.85\textwidth]{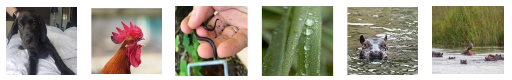}}
\caption{{Data overlap examples. We use Google Images to find similar or same images in ImageNetV2.}}
    \label{fig:data_overlap}
\end{figure*}

We provide similar image examples found by Google Images in Figure~\ref{fig:data_overlap}.

\section{CLIP-Auto Details}

\paragraph{Automated Prompt Generation} We describe our method to generate automated prompts in detail. Formally, the goal is to learn a prompt $\mathbf{x}_{\rm prompt} = z(\mathbf{x}_{\rm trig}, \mathbf{x}_{\rm label})$, where $z$ is the template such as ``\textspt{[T][T][T][T][C]}''. \textspt{[T]} indicates a trigger token, and \textspt{[C]} indicates the label text. The idea is to add a set of trigger tokens (i.e., \textspt{[T]}) to the label text according to the template. The process is framed as a prompt search task. As shown in Figure~\ref{fig:overview}, the trigger tokens are initialized as ``A photo of a'', then iteratively updated to minimize the classification loss over batches of training examples. At each search step, the change of loss corresponding to the replacement of a trigger token with another token in the vocabulary is computed by a first-order Taylor approximation~\cite{wallace-etal-2019-universal}. For each trigger token, we keep the top-$k$ candidate trigger tokens that lead to the smallest loss, formally: $ \mathcal{T}_{cand} = {\rm top}\mhyphen k_{t \in \mathcal{V}}[\mathcal{L}-\nabla_{\mathbf{t}}\mathcal{L}].$ $\mathcal{L}$ is the loss, $t$ is a candidate token from the vocabulary $\mathcal{V}$, $\mathbf{t}$ is the corresponding input embedding.
We evaluate the updated prompt at the current step and retain the prompt with the highest probability in the next step. In practice, we perform a left-to-right beam search over the top-$k$ candidate trigger tokens using the candidate sequences with the smallest loss at the current step. We use small beam sizes for efficiency consideration. For example, the trigger tokens converge to ``lovely picture show the'', which is used in the next step. The final prompt from the last step is returned. We use the ImageNet training set to find the prompts.

Rather than keeping the best candidate sequence of trigger tokens at each step as in the above setting, the best candidate from each step is concatenated and deduplicated. We select $n$ candidate sequences that lead to the best performance on a validation set sampled from ImageNet containing 10,000 images. Instead of ensembling over the probability space of multiple classifiers, we follow \cite{radford2021learning} to ensemble over the embedding space of the text. We use this as the default setting for CLIP-Auto. 

To keep the training as simple as possible, we use the same hyperparameters to search the automated prompts for all CLIP-Auto models. We use: top-$k$ equals 20 of the candidate trigger tokens; beam size equals 5 according to \cite{wallace-etal-2019-universal}; 1 GeForce GTX 1080 with a batch size of 512; the number of training steps equals 2000; each step takes approximately 3 minutes. We generate all CLIP-Auto models on the ImageNet training set and randomly choose 1,000 samples from the validation set to validate the best number of prompts. The number of automated prompts is: 49, 8, 186, 84, 7, 22 for ResNet50, ResNet101, ViT-B/32, ViT-B/16, ResNet50x4, and ResNet50x16 respectively.

\label{sec:listp}
\paragraph{Case Study} We show uncurated full sets of prompts for CLIP\footnote{\small{\url{https://github.com/openai/CLIP/blob/main/notebooks/Prompt_Engineering_for_ImageNet.ipynb}}\label{ft:clipprompt}} and CLIP-Auto in Table~\ref{tab:cliporig} to Table~\ref{tab:cliprn5016}. Automated prompts for each model are produced based on the same prompt template ``\textspt{[T][T][T][T][C].}''. We find that although the automated prompts are less interpretable, they are able to construct classifiers that predict comparably with the manual prompts, thanks to its flexibility in deriving customized prompts for each model. However, the impact of prompts synthesizing the classifiers is limited by the pre-trained image representations. Therefore we do not see a significant difference in the robustness performance.
\begin{table*}[]
 \centering
\begin{minipage}[t]{.3\linewidth}\vspace{0pt}
      \centering
      \resizebox{\textwidth}{!}{%
\begin{tabular}{l}
\toprule

a bad photo of a \{label\}  \\a photo of many \{label\}  \\a sculpture of a \{label\}  \\a photo of the hard to see \{label\}  \\a low resolution photo of the \{label\}  \\a rendering of a \{label\}  \\graffiti of a \{label\}  \\a bad photo of the \{label\}  \\a cropped photo of the \{label\}  \\a tattoo of a \{label\}  \\the embroidered \{label\}  \\a photo of a hard to see \{label\}  \\a bright photo of a \{label\}  \\a photo of a clean \{label\}  \\a photo of a dirty \{label\}  \\a dark photo of the \{label\}  \\a drawing of a \{label\}  \\a photo of my \{label\}  \\the plastic \{label\}  \\a photo of the cool \{label\}  \\a close-up photo of a \{label\}  \\a black and white photo of the \{label\}  \\a plastic \{label\}  \\
a photo of the small \{label\}  \\a photo of the weird \{label\}  \\a bright photo of the \{label\}  \\a cropped photo of a \{label\}  \\ a photo of the large \{label\}  \\ \bottomrule
\end{tabular}%
}
\end{minipage}
\hspace{0.1in}
\begin{minipage}[t]{.3\linewidth}\vspace{0pt}
      \centering
      \resizebox{\textwidth}{!}{%
\begin{tabular}{l}
\toprule

the \{label\} in a video game \\ a sketch of a \{label\}  \\a doodle of the \{label\}  \\a origami \{label\}  \\a low resolution photo of a \{label\}  \\the toy \{label\}  \\a rendition of the \{label\}  \\a photo of the clean \{label\}  \\a photo of a large \{label\}  \\a rendition of a \{label\}  \\a photo of a nice \{label\}  \\a photo of a weird \{label\}  \\a blurry photo of a \{label\}  \\a cartoon \{label\}  \\art of a \{label\}  \\a sketch of the \{label\}  \\a embroidered \{label\}  \\a pixelated photo of a \{label\}  \\itap of the \{label\}  \\a jpeg corrupted photo of the \{label\}  \\a good photo of a \{label\}  \\a plushie \{label\}  \\a photo of the nice \{label\}  \\the cartoon \{label\}  \\art of the \{label\}  \\a drawing of the \{label\}  \\ a black and white photo of a \{label\}  \\ \bottomrule
\end{tabular}%
}
\end{minipage}
\begin{minipage}[t]{.3\linewidth}\vspace{0pt}
      \centering
      \resizebox{\textwidth}{!}{%
\begin{tabular}{l}
\toprule

the plushie \{label\}  \\a dark photo of a \{label\}  \\itap of a \{label\}  \\graffiti of the \{label\}  \\a toy \{label\}  \\itap of my \{label\}  \\a photo of a cool \{label\}  \\a photo of a small \{label\}  \\a tattoo of the \{label\} \\
a photo of the dirty \{label\}  \\a jpeg corrupted photo of a \{label\}  \\a blurry photo of the \{label\}  \\a photo of the \{label\}  \\a good photo of the \{label\}  \\a rendering of the \{label\}  \\a \{label\} in a video game \\ a photo of one \{label\}  \\a doodle of a \{label\}  \\a close-up photo of the \{label\}  \\a photo of a \{label\}  \\the origami \{label\}  \\
a painting of the \{label\}  \\a painting of a \{label\}  \\a pixelated photo of the \{label\}  \\a sculpture of the \{label\}  \\a photo of the large \{label\}  \\ 

\bottomrule
\end{tabular}
}

      \end{minipage}
\caption{{ Uncurated prompts of CLIP}.}
\label{tab:cliporig}

\end{table*}

\begin{table*}[]
 \centering
\begin{minipage}[t]{.3\linewidth}\vspace{0pt}
      \centering
      \resizebox{\textwidth}{!}{%
\begin{tabular}{l}
\toprule

bri dexter suppose an \{label\}  \\ 
amazingly wri means an \{label\}  \\ 
annually allegedly {]} a \{label\}  \\ 
instead :{]} frifotos an \{label\}  \\ 
atively factfriday !.. an \{label\}  \\ 
instead aper frifotos an \{label\}  \\ 
also .\# pics numerous \{label\}  \\
atively factfriday an \{label\}  \\ 
ever cheerful about an \{label\}  \\ 
also intre acquainted an \{label\}  \\ 
ij favourites factfriday a \{label\}  \\
unpopular typically {]} a \{label\}  \\
annually ates {]} typical \{label\}  \\
annually ates {]} this \{label\}  \\ 
bi ant thousands interesting \{label\}  \\ 
instead continuation frifotos an \{label\}  \\

instead glorious frifotos an \{label\}  \\ 

esper sees !). googled \{label\}   \\\bottomrule
\end{tabular}%
}
\end{minipage}
\hspace{0.1in}
\begin{minipage}[t]{.3\linewidth}\vspace{0pt}
      \centering
      \resizebox{\textwidth}{!}{%
\begin{tabular}{l}
\toprule

annually ates {]} interesting \{label\}  \\ 
amazingly fascin about an \{label\}  \\
instead cro mesmerizing an \{label\}  \\ 
amazingly sth discover an \{label\}  \\ 
( talking about an \{label\}  \\ 
affection randomly about an \{label\}  \\ 
complete fascin about an \{label\}  \\ 
also goog acquainted an \{label\}  \\ 
ever fascin about an \{label\}  \\ 
although photo of a \{label\}  \\ 
rarely hein suppose an \{label\}  \\ 
ever genus admire an \{label\}  \\ 
rarely exc suppose an \{label\}  \\ 
esper rained !). googled \{label\}  \\
crazy factfriday .) an \{label\}  \\ 
rarely easter suppose an \{label\}  \\  \bottomrule
\end{tabular}%
}
\end{minipage}
\begin{minipage}[t]{.3\linewidth}\vspace{0pt}
      \centering
      \resizebox{\textwidth}{!}{%
\begin{tabular}{l}
\toprule

potd enjo about an \{label\}  \\
among talking about an \{label\}  \\
amazingly uni tically an \{label\}  \\
ably prett mous an \{label\}  \\ 
incredibly fascin about an \{label\}  \\
singul thing factfriday a \{label\}  \\ 
ingh random )... an \{label\}  \\ 
commonly atio {]}: a \{label\}  \\ 
atively {]} awesome \{label\}  \\ 
hetero tional " beautiful \{label\}  \\ 
tious query !). googled \{label\}  \\ 
habit rained !). googled \{label\}  \\ 
besides only !). an \{label\}  \\ 
ever thing interesting an \{label\}  \\ 
wonderfully kan )... an \{label\}  \\ 
\bottomrule
\end{tabular}
}

      \end{minipage}
\caption{{ Uncurated prompts of CLIP-Auto of ResNet50}.}
\end{table*}


\begin{table*}[]
 \centering
\begin{minipage}[t]{.3\linewidth}\vspace{0pt}
      \centering
      \resizebox{\textwidth}{!}{%
\begin{tabular}{l}
\toprule

awesome                     coo            ing                led          \{label\} \\
and                         awesome        ==\textgreater{}   a            \{label\} \\
classi                      ele            primarily          smaller      \{label\} \\
awesome                     hob            ing                led          \{label\} \\
awesome                     wonderful      thero              led          \{label\} \\
brightly                    ative          -                  like         \{label\} \\
jou                         photo          of                 a            \{label\} \\
blu                         related        typical            smaller      \{label\} \\
brightly                    ative          spo                old          \{label\} \\
awesome                     very           ver                oldest       \{label\} \\
super                       awesome        neat               a            \{label\} \\
awesome                     lovely         ca                 fiable       \{label\} \\
especially                  also           !).                smaller      \{label\} \\
contributed                 potty          ==\textgreater{}   ordinary     \{label\} \\
.....\#                     trivia         !:                 smaller      \{label\} \\
potentially                 awesome        -                  a            \{label\} \\
funfactfriday               awesome        ==\textgreater{}   a            \{label\} \\
contributed                            ==\textgreater{}   typical      \{label\} \\
                                  ==\textgreater{}   large        \{label\} \\
unto                        picoftheday    !).                smaller      \{label\} \\
awesome                     (!)            ()                ous          \{label\} \\
relatively                  awesome        !                  a            \{label\} \\
potentially                 awesome        !).                a            \{label\} \\
,\#                         awesome        neat               a            \{label\} \\
contributed                            ==\textgreater{}   ordinary     \{label\} \\
.....\#                     trivia         ..                 smaller      \{label\} \\
theworld                    ness           !:                 smaller      \{label\} \\
:                          ...:           \textbackslash{}'  interesting  \{label\} \\
supposedly                  hob            ing                led          \{label\} \\
awesome                     gin            ing                led          \{label\} \\
large                       two            -                  covered      \{label\} \\
exac                        latin          -                  covered      \{label\} \\
funfactfriday               omg            !).                a            \{label\} \\
quiz                        namesake       )                  smaller      \{label\} \\
commonly                    someday        !).                smaller      \{label\} \\
of                          picoftheday    ..                 smaller      \{label\} \\
awesome                     ...:           approximately      smaller      \{label\} \\
awesome                     :-             ca                 formed       \{label\} \\
awesome                     neat           an                 ering        \{label\} \\
awesome                     dy             an                 ering        \{label\} \\
awesome                     wonderful      ing                led          \{label\} \\
awesome                     (!)            wonderful          ous          \{label\} \\
larger                      ative          -                  like         \{label\} \\
seriously                   photos         of                 a            \{label\} \\
relatively                  photos         of                 a            \{label\} \\
ously                       awesome        !                  a            \{label\} \\
funfactfriday               awesome        !).                a            \{label\} \\
amazingly                   awesome        neat               a            \{label\} \\
smaller                     awesome        neat               a            \{label\} \\
contributed                            ==\textgreater{}   favorite     \{label\} \\
contributed                            ==\textgreater{}   ordinary     \{label\} \\
contributed                           ==\textgreater{}   nice         \{label\} \\
contributed                 potty          ==\textgreater{}   those        \{label\} \\
naturally                   related        )                  smaller      \{label\} \\
of                          picoftheday    !).                smaller      \{label\} \\
theworld                    fascin         !:                 smaller      \{label\} \\
theworld                    indication     !:                 smaller      \{label\} \\
classi                      closely        !:                 smaller      \{label\} \\
                      ...:                          interesting  \{label\} \\
awesome                     ...:           share              smaller      \{label\} \\ \bottomrule
\end{tabular}%
}

\end{minipage}
\hspace{0.1in}
\begin{minipage}[t]{.3\linewidth}\vspace{0pt}
      \centering
      \resizebox{\textwidth}{!}{%
\begin{tabular}{l}
\toprule

seated                      photo          of                 a            \{label\} \\
sson                        photo          of                 a            \{label\} \\
relatively                  pictures       !                  a            \{label\} \\
sively                      awesome        !                  a            \{label\} \\
exceptionally               awesome        !                  a            \{label\} \\
potentially                 awesome        !                  a            \{label\} \\
things                      awesome        !).                a            \{label\} \\
funfactfriday               hooray         !).                a            \{label\} \\
or                          awesome        !).                a            \{label\} \\
aside                       funfactfriday  !).                a            \{label\} \\
omfg                        funfactfriday  !).                a            \{label\} \\
just                        awesome        neat               a            \{label\} \\
supposedly                  awesome        )                  a            \{label\} \\
kidding                     awesome        )                  a            \{label\} \\
controversial               awesome        )                  a            \{label\} \\
funfactfriday               awesome        :                  a            \{label\} \\
obligatory                  awesome        ==\textgreater{}   a            \{label\} \\
contributed                           ==\textgreater{}   large        \{label\} \\
contributed                           ==\textgreater{}   which        \{label\} \\
contributed                            ==\textgreater{}   unwanted     \{label\} \\
contributed                          ==\textgreater{}   empty        \{label\} \\
contributed                 potty          ==\textgreater{}   empty        \{label\} \\
contributed                 potty          ==\textgreater{}   wonderful    \{label\} \\
contributed                          ==\textgreater{}   those        \{label\} \\
odd                         inspired       ==\textgreater{}   typical      \{label\} \\
presents                    inspired       ==\textgreater{}   typical      \{label\} \\
----------------            inspired       ==\textgreater{}   typical      \{label\} \\
----------------            ij             ==\textgreater{}   smaller      \{label\} \\
----------------            ij             favourite          smaller      \{label\} \\
neh                         informative    favourite          smaller      \{label\} \\
pas                         related        peoples            smaller      \{label\} \\
:                           related        typical            smaller      \{label\} \\
:                           related        )                  smaller      \{label\} \\
highly                      namesake       )                  smaller      \{label\} \\
often                       namesake       )                  smaller      \{label\} \\
often                       similar        !).                smaller      \{label\} \\
often                       separately     !).                smaller      \{label\} \\
ima                         separately     !).                smaller      \{label\} \\
formally                    same           !).                smaller      \{label\} \\
commonly                    holidays       !).                smaller      \{label\} \\
related                     --             !).                smaller      \{label\} \\
grand                       --             !).                smaller      \{label\} \\
better                      --             !).                smaller      \{label\} \\
previous                    --             !).                smaller      \{label\} \\
taller                      lished         !).                smaller      \{label\} \\
taller                      spoiled        !).                smaller      \{label\} \\
taller                      challenged     !).                smaller      \{label\} \\
preten                      broader        !).                smaller      \{label\} \\
preten                      later          !).                smaller      \{label\} \\
photo                       picoftheday    !).                smaller      \{label\} \\
of                          picoftheday                     smaller      \{label\} \\
\_\_\_\_\_                  random         ..                 smaller      \{label\} \\
.....\#                     trivia         toftheday          smaller      \{label\} \\
.....\#                     fascin         !:                 smaller      \{label\} \\
!)                          fascin         !:                 smaller      \{label\} \\
classi                      symbolic       !:                 smaller      \{label\} \\
classi                      introduced     !:                 smaller      \{label\} \\

awww                        impressive     ca                 fiable       \{label\} \\ \bottomrule
\end{tabular}%
}

\end{minipage}
\begin{minipage}[t]{.3\linewidth}\vspace{0pt}
      \centering
      \resizebox{\textwidth}{!}{%
\begin{tabular}{l}
\toprule

lovel                       ...:           share              interesting  \{label\} \\
hamp                        ...:           share              famous       \{label\} \\
classi                      inged          purposes           smaller      \{label\} \\
dana                        ...:           ordinary           large        \{label\} \\
...\#                       ...:           attractive         large        \{label\} \\
awesome                     lovely         ca                 eled         \{label\} \\
awesome                     lovely         ca                 oldest       \{label\} \\
awesome                     lovely         ini                oldest       \{label\} \\
awesome                     gently         be                 oldest       \{label\} \\
awesome                     seemingly      ver                oldest       \{label\} \\
awesome                     neat           ver                oldest       \{label\} \\
awesome                     european       an                 nicest       \{label\} \\
awesome                     bic            an                 nicest       \{label\} \\
awesome                     impressive     an                 yed          \{label\} \\
awesome                     cro            an                 other        \{label\} \\
awesome                     cro            ste                other        \{label\} \\
awesome                     hob            uni                led          \{label\} \\
amazingly                   hob            ing                led          \{label\} \\
awesome                     delightful     ing                led          \{label\} \\
awesome                     cool           ing                led          \{label\} \\
awesome                     wonderful      thero              important    \{label\} \\
awesome                     funfactfriday  actual             ous          \{label\} \\
awesome                     funfactfriday  absolutely         ous          \{label\} \\
small                       (!)            some               huge         \{label\} \\
small                       ulously        |                  huge         \{label\} \\
small                       ative          |                  huge         \{label\} \\
brightly                    ative          -                  smelly       \{label\} \\
thest                       ative          -                  gorgeous     \{label\} \\
several                     ative          -                  esque        \{label\} \\
large                       ative          -                  esque        \{label\} \\
large                       single         -                  esque        \{label\} \\
large                       pra            -                  covered      \{label\} \\
distinctive                 backward       -                  covered      \{label\} \\
registered                  pee            -                  covered      \{label\} \\
commonly                    reasonable     -                  covered      \{label\} \\
registered                  latin          -                  covered      \{label\} \\
and                         latin          -                  covered      \{label\} \\
exce                        dual           -                  covered      \{label\} \\
common                      continuous     -                  covered      \{label\} \\
common                      µ             -                  sized        \{label\} \\
common                      ities          -                  sized        \{label\} \\
recre                       ities          -                  sized        \{label\} \\
recre                       wanna          -                  sized        \{label\} \\ 
awesome                     macro          ca                 fiable       \{label\} \\
awesome                     hob            tr                 led          \{label\} \\
awesome                     favourite      ing                led          \{label\} \\
thest                       ative          -                  like         \{label\} \\
ari                         ly             -                  covered      \{label\} \\
common                      tively         -                  covered      \{label\} \\
:@                          ...:           ¥                interesting  \{label\} \\
share                       ...:           ¥                interesting  \{label\} \\
theo                        ...:           /                  interesting  \{label\} \\
aerop                       ...:           share              interesting  \{label\} \\
hamp                        ...:           share              large        \{label\} \\
...: share large \{label\}  ...:           share              japanese     \{label\} \\
awesome                     ...:           out                smaller      \{label\} \\
awesome                     ...:           largely            smaller      \{label\} \\
awesome                     ...:           ca                 other        \{label\} \\
awesome                     ...:           ca                 formed       \{label\} \\
awesome                     thero          ca                 funded       \{label\} \\
beautiful                   Ê              ca                 funded       \{label\} \\
gorgeous                    Ê              ca                 funded       \{label\} \\
gorgeous                                ca                 funded       \{label\} \\
...\#                       ...:           awesome            interesting  \{label\} \\
awesome                     Ê              ca                 funded       \{label\} \\
...\#                       ...:           noisy              interesting  \{label\} \\
...\#                       ...:           ordinary           large        \{label\} \\
\bottomrule
\end{tabular}
}

      \end{minipage}
\caption{{ Uncurated prompts of CLIP-Auto of ViT-B/32.}}
\end{table*}

\begin{table*}[]
 \centering
\begin{minipage}[t]{.45\linewidth}\vspace{0pt}
      \centering
      \resizebox{\textwidth}{!}{%
\begin{tabular}{l}
\toprule

various unlike ": a \{label\} \\ til photo -- a \{label\} \\ unpopular introduce .âĢ¦ a \{label\} \\ completely familiar recognizable wonderful \{label\} \\ til correctly -- a \{label\} \\ many other recognizable an \{label\} \\ adop circulating recognizable entire \{label\} \\ awesome orient ce sized \{label\} \\ \bottomrule
\end{tabular}%
}
\caption{{ Uncurated prompts of CLIP-Auto of ResNet101.}}
\end{minipage}
\hspace{0.1in}
\begin{minipage}[t]{.4\linewidth}\vspace{0pt}
      \centering
      \resizebox{\textwidth}{!}{%
\begin{tabular}{l}
\toprule

wednesday goo explanations an \{label\} \\ sforsale fineartamerica nbd about \{label\} \\ neva awesome didyouknow a \{label\} \\ gee friend didyouknow a \{label\} \\ aun holidays didyouknow a \{label\} \\ kic holidays didyouknow a \{label\} \\ eco coalition -- delightful \{label\} \\ \bottomrule
\end{tabular}%
}
\caption{{ Uncurated prompts of CLIP-Auto of ResNet50x4.}}
\end{minipage}
\end{table*}

\begin{table*}[]
 \centering
\begin{minipage}[t]{.3\linewidth}\vspace{0pt}
      \centering
      \resizebox{\textwidth}{!}{%
\begin{tabular}{l}
\toprule

then coolest )… a \{label\} \\
because unusual )… a \{label\} \\
neat awsome ........... a \{label\} \\
heavily etzinteresting a \{label\} \\
what awsome -- a \{label\} \\
considered awesome :) a \{label\} \\
awesomeness !), )… a \{label\} \\
........awsome -- a \{label\} \\
awesomeness perhaps )… a \{label\} \\
totally !), )… a \{label\} \\
a typical inviting a \{label\} \\
ably awesom... a \{label\} \\
wonderful vely interesting a \{label\} \\
fascinating renowned :) a \{label\} \\
how awsome ) a \{label\} \\
randomly !), )… a \{label\} \\
because fascinating )… a \{label\} \\
then awesomeness )… a \{label\} \\
wonderful fortunately )… a \{label\} \\
precisely awesomeness :" a \{label\} \\
a displainviting a \{label\} \\
ably awesomsure a \{label\} \\
what awsome >> a \{label\} \\
fairly interesting !!!!!! a \{label\} \\
because hooray )… a \{label\} \\
)… awesomeness :) a \{label\} \\
arbitrparticularly gosh a \{label\} \\

\bottomrule
\end{tabular}%
}

\end{minipage}
\hspace{0.1in}
\begin{minipage}[t]{.3\linewidth}\vspace{0pt}
      \centering
      \resizebox{\textwidth}{!}{%
\begin{tabular}{l}
\toprule

thatsmoly :) a \{label\} \\
thatsjust :) a \{label\} \\
because neat )… a \{label\} \\
then craziest )… a \{label\} \\
then fascinating :" a \{label\} \\
)… awesomrecognizable a \{label\} \\
awesome wonderfully )( a \{label\} \\
longest etzinteresting a \{label\} \\
neat :)) amazing a \{label\} \\
awesomeness fer :) a \{label\} \\
what awsome )… a \{label\} \\
fascinating snazzy ).. a \{label\} \\
supposedly awesome :) a \{label\} \\
fascinating ilove :) a \{label\} \\
thatswanted :) a \{label\} \\
considered fascinating :) this \{label\} \\
awesomeness fascinating :) a \{label\} \\
fascinating jst ,) a \{label\} \\
because awesome)… a \{label\} \\
because fyi )… a \{label\} \\
then fancy )… a \{label\} \\
individually awesomeness :" a \{label\} \\
a photo of a \{label\} \\
a frontal bestoa \{label\} \\
a rentbestoa \{label\} \\
a industrialbestoa \{label\} \\
a typical bestoa \{label\} \\

\bottomrule
\end{tabular}%
}
\end{minipage}
\begin{minipage}[t]{.3\linewidth}\vspace{0pt}
      \centering
      \resizebox{\textwidth}{!}{%
\begin{tabular}{l}
\toprule

deeply bilateral inviting a \{label\} \\
recognizable particularly ugliest a \{label\} \\

arbitractual oooooo a \{label\} \\
incredibly actual oooooo a \{label\} \\
incredibly unusual  a \{label\} \\
incredibly behold  a \{label\} \\
incredibly behold :-) a \{label\} \\
)… behold :-) a \{label\} \\
)… behold awesomeness a \{label\} \\
)… behold exciting a \{label\} \\
ably awesomfascinating a \{label\} \\
ably awesominstance a \{label\} \\
cool wonderfully )( a \{label\} \\
instantly awsome admire a \{label\} \\
thus awsome admire a \{label\} \\
incredibly awsome ) a \{label\} \\
incredibly unusual oooooo a \{label\} \\
wonderful etzinteresting a \{label\} \\
coolest justsaying :). a \{label\} \\
awesomeaping :) a \{label\} \\
looodesperately admire a \{label\} \\
thus awsome beautiful a \{label\} \\
actually awsome -- a \{label\} \\
fairly interesting ).. a \{label\} \\
wow awesomeness why a \{label\} \\
a demoinviting a \{label\} \\
the excinviting a \{label\} \\
compare recognizable inviting a \{label\} \\
because awesomeness )… a \{label\} \\
wonderful  ! a \{label\} \\
\bottomrule
\end{tabular}
}
      \end{minipage}
\caption{{ Uncurated prompts of CLIP-Auto of ViT-B/16.}}
\end{table*}

\begin{table*}[]
 \centering
\begin{minipage}[t]{.3\linewidth}\vspace{0pt}
      \centering
      \resizebox{\textwidth}{!}{%
\begin{tabular}{l}
\toprule

enigffect found '\# \{label\} \\
\#:'topics another \{label\} \\
tessenulogically cool \{label\} \\
enigffect found '\# \{label\} \\
\#:'topics another \{label\} \\
tessenulogically cool \{label\} \\
small freaking spegir \{label\} \\

\bottomrule
\end{tabular}%
}
\end{minipage}
\begin{minipage}[t]{.3\linewidth}\vspace{0pt}
      \centering
      \resizebox{\textwidth}{!}{%
\begin{tabular}{l}
\toprule

whoa kineusing a \{label\} \\
expectthageneric impressive \{label\} \\
pak,… random impressive \{label\} \\
dingh?) lovely \{label\} \\
quite famili...!!! sized \{label\} \\
inely similarities – sized \{label\} \\
archelic :-) colorful \{label\} \\

yay perfectly !), amazing \{label\} \\
\_\_\_\_ fact gratextinct \{label\} \\
\bottomrule
\end{tabular}%
}
\label{tab:clippvitb16}
\end{minipage}
\begin{minipage}[t]{.3\linewidth}\vspace{0pt}
      \centering
      \resizebox{\textwidth}{!}{%
\begin{tabular}{l}
\toprule

umtom word : \{label\} \\
undant j) terrifying \{label\} \\
those kalically amazing \{label\} \\
spare sovereignically actual \{label\} \\
everydayphoto unmatched a \{label\} \\
oooo-- coolest a \{label\} \\
pretty "\# coolest a \{label\} \\
simultaneously almost coolest a \{label\} \\
weird ´ radinvolving \{label\} \\

\bottomrule
\end{tabular}
}

      \end{minipage}
\caption{{ Uncurated prompts of CLIP-Auto of ResNet50x16.}}
\label{tab:cliprn5016}
\end{table*}

\section{Example Images of Distribution Shifts}
\begin{figure*}
\centering
\subcaptionbox{{ Seven natural distribution shifts. Examples are from ImageNetV2, ImageNet-A, ImageNet-R, ImageNet Sketch, ImageNet-Vid, ObjectNet, and Youtube-BB.}\label{fig:naturalexamples}}{\includegraphics[width=0.9\textwidth, height=0.65\textwidth]{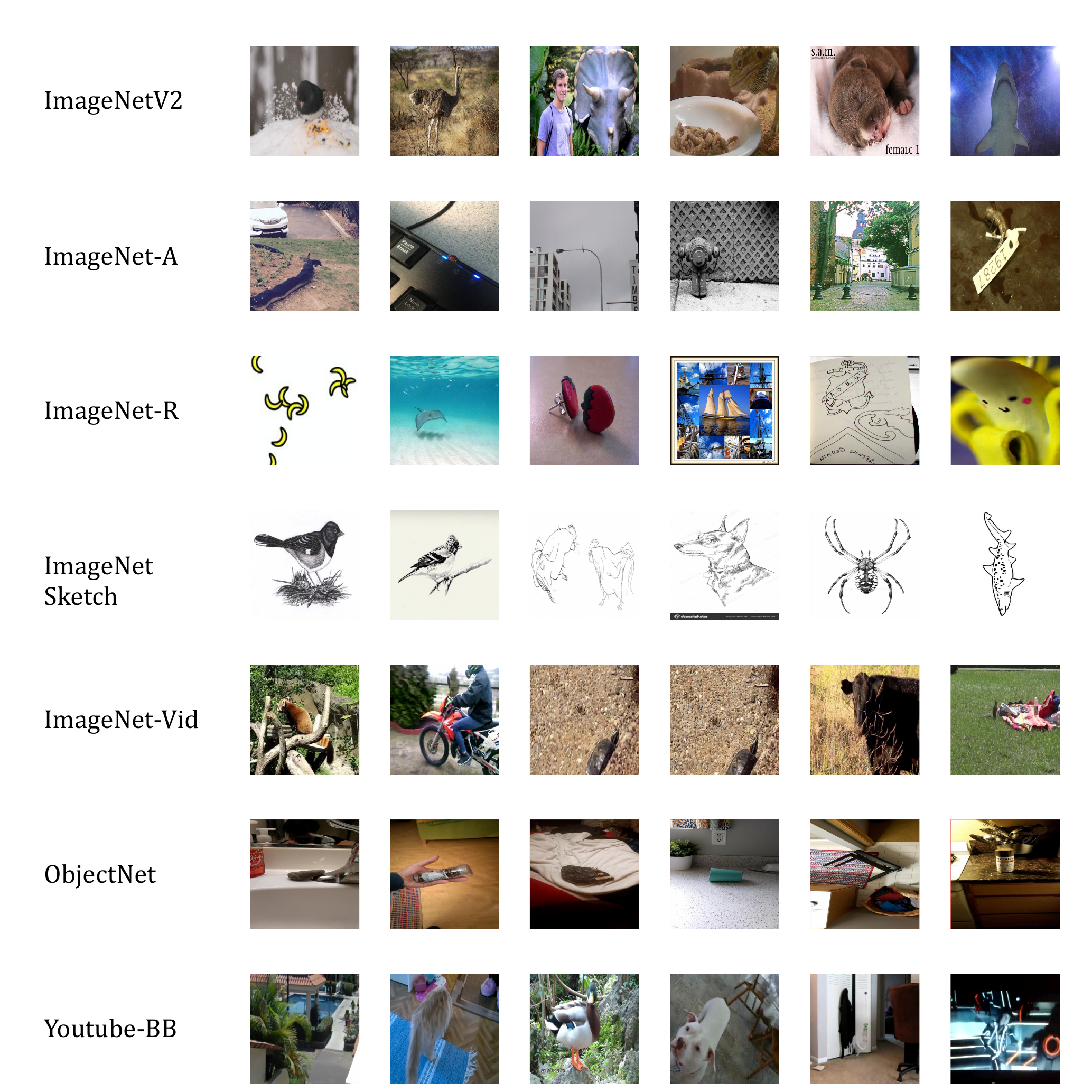}} \\
\subcaptionbox{{ Three synthetic distribution shifts. Examples are from ImageNet-C, video frames of ImageNet-P, and Stylized ImageNet.}\label{fig:syntheticexamples}}{\includegraphics[width=0.9\textwidth,height=0.55\textwidth]{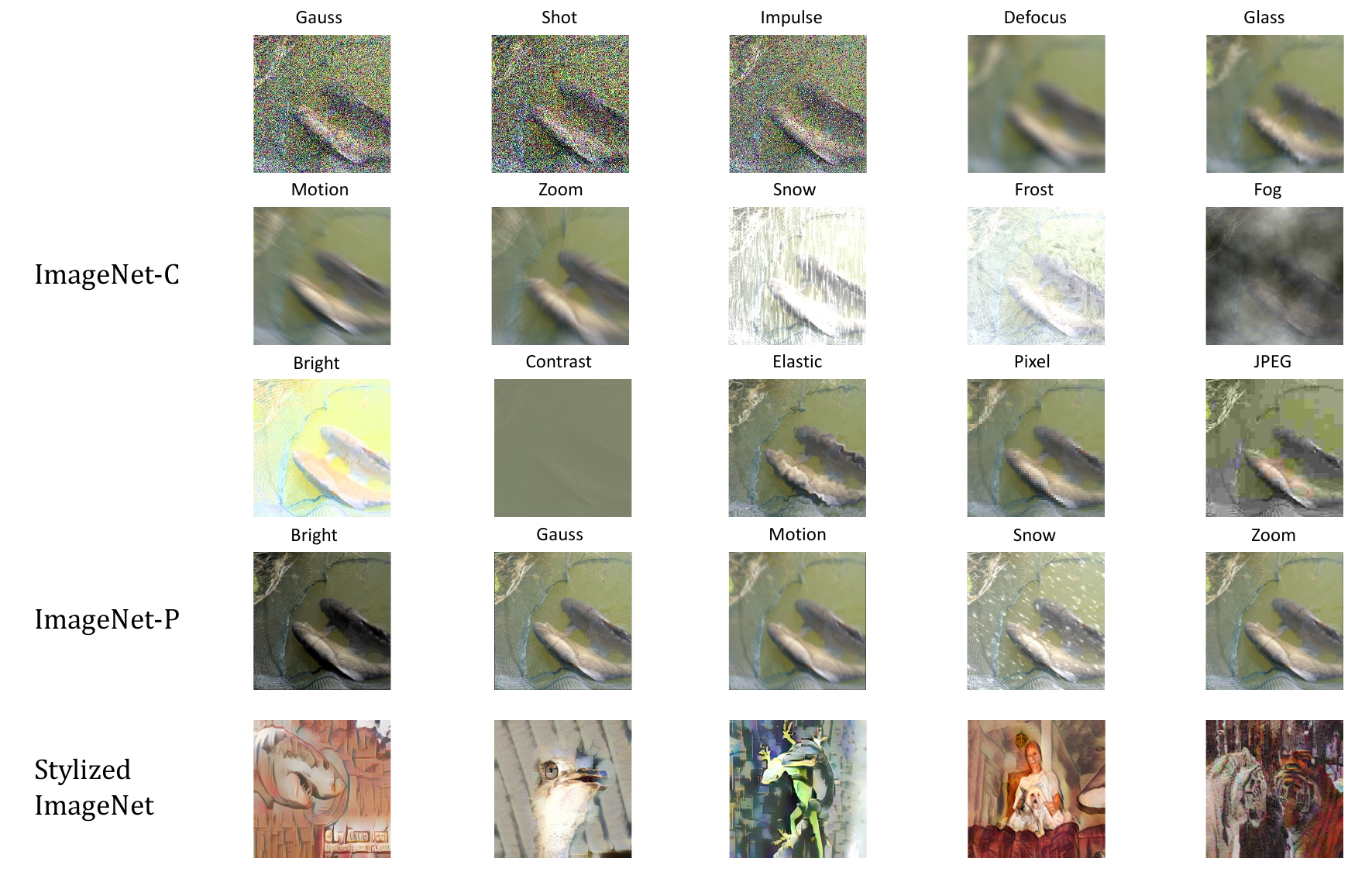}}
\caption{{ Distribution shift images.}}
    \label{fig:distributionshiftexample}
\end{figure*}

We show example images of all distribution shifts in our \benchmark\ benchmark in Figure~\ref{fig:distributionshiftexample}.

\section{Additional Related Work}
Natural language prompt engineering is useful for zero (or few)-shot performance in NLP, such as large language models~\cite{wang2020language,wang-etal-2022-deepstruct} including GPT-3~\cite{brown2020language}, GPT-4~\cite{gpt4}, and Gemini~\cite{team2023gemini}. Especially, the manually created prompts~\cite{petroni2020context} are used for fact retrieval. While the primary usage of the natural language prompts aims to improve the performance of NLP tasks, we aim to use the natural language prompts to improve the robustness of performance in other domains, e.g., image classification. The prompt learning methods~\cite{zhang2020contrastive,zhou2022conditional} employ the continuous prompt method, which is less explainable. As concluded in Sec.~\ref{sec:dataoverlap}, the dominant factor is the pre-trained image representation. Therefore, similar to our method, these prompt learning approaches mainly improve the efficiency of training (avoiding updating all model parameters) instead of robustness enhancement.

\end{document}